\documentclass[journal]{IEEEtran}
\usepackage{cite}
\ifCLASSINFOpdf
\else
\fi
\usepackage{amssymb,amsmath,amsthm,graphicx,psfrag,cite}
\usepackage[usenames]{color}
\usepackage{epsfig}
\usepackage{subfigure}
\usepackage{hyperref}
\hypersetup{
    bookmarks=false,         
    pdftitle={Compressive Imaging using Approximate Message Passing and a Markov-Tree Prior},    
    pdfauthor={Subhojit Som and Philip Schniter},     
    colorlinks,%
    citecolor=black,%
    filecolor=black,%
    linkcolor=black,%
    urlcolor=black
}
 \newcommand{\defn}{\triangleq}

 \newcommand{\hvec}[1]{\ensuremath{\Hat{\boldsymbol{#1}}}}
 \renewcommand{\vec}[1]{\ensuremath{\boldsymbol{#1}}}

 \newcommand{\norm}[1]{\ensuremath{\| #1 \|}}
 \newcommand{\mc}[1]{\ensuremath{\mathcal{#1}}}

 \newcommand{\Real}{{\mathbb{R}}}

 \newcommand{\of}[1]{^{\scriptscriptstyle (#1)}}
 \renewcommand{\sp}[1]{\ensuremath{\mathcal{#1}}}

 \DeclareMathOperator{\E}{E}

 \renewcommand{\eqref}[1]{(\ref{eq:#1})}

 \newcommand{\figref}[1]{Fig.~\ref{fig:#1}}
 
 \newcommand{\secref}[1]{Section~\ref{sec:#1}}

\newcommand{\ve}[1]{\ensuremath{\boldsymbol{#1}}}
\renewcommand{\sp}[1]{\ensuremath{\mathcal{#1}}}
\renewcommand{\eqref}[1]{(\ref{eqn:#1})}
\newcommand{\giv}{\,|\,}

\newcommand{\nL}{_{n,\text{\sf L}}}
\newcommand{\nS}{_{n,\text{\sf S}}}
\newcommand{\jL}{_{j,\text{\sf L}}}
\newcommand{\jS}{_{j,\text{\sf S}}}
\begin{document}
\setlength{\arraycolsep}{0.8mm}
%
\title{Compressive Imaging using Approximate Message Passing and a Markov-Tree Prior}
\author{Subhojit Som\IEEEauthorrefmark{1} and 
	Philip Schniter\IEEEauthorrefmark{2}
	\thanks{\IEEEauthorrefmark{1}Dr. Subhojit Som is with the School of Electrical and Computer Engineering, Georgia Institute of Technology, 75 Fifth St NW, Atlanta, GA 30308, email: subhojit@gatech.edu, phone 404.894.2901, fax 404.894.8363.}
	\thanks{\IEEEauthorrefmark{2}Prof. Philip Schniter is with the Department of Electrical and Computer Engineering, The Ohio State University, 2015 Neil Ave., Columbus, OH 43202, email: schniter@ece.osu.edu, phone 614.247.6488, fax 614.292.7596.}
	\thanks{This work was supported in part by NSF grant CCF-1018368, AFOSR grant FA9550-06-1-0324, DARPA/ONR grant N66001-10-1-4090, and an allocation of computing time from the Ohio Supercomputer Center.}
	\thanks{This work was presented in part at the 2010 Asilomar Conference on Signals, Systems, and Computers \cite{Som:ASIL:10}.}
	\thanks{Please direct all correspondence to Philip Schniter at the above address.}
}
\maketitle
\begin{abstract}
We propose a novel algorithm for compressive imaging that exploits both
the sparsity and persistence across scales found in the 2D wavelet 
transform coefficients of natural images.
Like other recent works, we model wavelet structure using a hidden Markov tree (HMT)
but, unlike other works, ours is based on loopy belief propagation (LBP).
For LBP, we adopt a recently proposed ``turbo'' message passing schedule 
that alternates between exploitation of HMT structure and
exploitation of compressive-measurement structure.
For the latter, we leverage Donoho, Maleki, and Montanari's recently proposed 
approximate message passing (AMP) algorithm.
Experiments with a large image database suggest that, relative to existing 
schemes, our turbo LBP approach yields state-of-the-art reconstruction 
performance with substantial reduction in complexity.
\end{abstract}
\section{Introduction}				\label{sec:intro}
In compressive imaging \cite{Romberg:SPM:08}, we aim to estimate an image 
$\ve{x} \in \mathbb{R}^N$ from $M\leq N$ noisy linear observations $\ve{y} \in \mathbb{R}^M$,
\begin{eqnarray}
\ve{y} 
&=& \ve{\Phi x} + \ve{w} 
\,=\, \ve{\Phi  \Psi \theta} + \ve{w},  \label{eqn:system}
\end{eqnarray}
assuming that the image has a representation $\ve{\theta}\in\Real^N$ in some wavelet
basis~$\ve{\Psi}$ (i.e., $\ve{x}=\ve{\Psi \theta}$) containing only a few ($K$) 
large coefficients (i.e., $K \ll N$).
In \eqref{system}, $\ve{\Phi} \in \mathbb{R}^{M \times N}$ is a known measurement matrix and 
$\ve{w} \sim \sp{N}(\ve{0}, \sigma^2\ve{I})$ is additive white Gaussian noise.
Though $M\!<\!N$ makes the problem ill-posed, it has been shown that $\ve{x}$ can be 
recovered from $\ve{y}$ when $K$ is adequately small and $\ve{\Phi}$ is incoherent with 
$\ve{\Psi}$ \cite{Romberg:SPM:08}.
The wavelet coefficients of natural images are known to have an additional structure 
known as {\em persistence across scales} (PAS) \cite{Mallat:Book:08}, which we now describe. 
For 2D images, the wavelet coefficients are naturally organized into quad-trees, where
each coefficient at level $j$ acts as a parent for four child coefficients at level $j\!+\!1$.
The PAS property says that, if a parent is very small, then all of its 
children are likely to be very small;
similarly, if a parent is large, then it is likely that some (but not necessarily all) of its children will also be large.

Several authors have exploited the PAS property for compressive imaging 
\cite{Baraniuk:TIT:10,Duarte:ICASSP:08,HeCarin:TSP:09,HeCarin:SPL:10}.
The so-called ``model-based'' approach \cite{Baraniuk:TIT:10} is a deterministic 
incarnation of PAS that leverages a restricted union-of-subspaces and manifests as 
a modified CoSaMP \cite{Needell:ACHA:09} algorithm.
Most approaches are Bayesian in nature, exploiting the fact that 
PAS is readily modeled by a \emph{hidden Markov tree} (HMT) \cite{Crouse:TSP:98}.
The first work in this direction appears to be \cite{Duarte:ICASSP:08},
where an iteratively re-weighted $\ell_1$ algorithm, generating
an estimate of $\vec{x}$, was alternated with a Viterbi algorithm, generating
an estimate of the HMT states.
More recently, HMT-based compressive imaging has been attacked using modern Bayesian 
tools \cite{Robert:Book:04}.
For example, \cite{HeCarin:TSP:09} used Markov-chain Monte-Carlo (MCMC), 
which is known to yield correct posteriors after convergence.
For practical image sizes, however, convergence takes an impractically long time, 
and so MCMC must be terminated early, at which point its performance may suffer.
Variational Bayes (VB) can sometimes offer a better performance/complexity
tradeoff, motivating the approach in \cite{HeCarin:SPL:10}.
Our experiments indicate that, while \cite{HeCarin:SPL:10} indeed offers 
a good performance/complexity tradeoff, it is possible to do significantly better.

In this paper, we propose a novel approach to HMT-based compressive imaging
based on loopy belief propagation \cite{Frey:ANIPS:98}.
For this, we model the coefficients in $\vec{\theta}$ as conditionally Gaussian 
with variances that depend on the values of HMT states, and we propagate 
beliefs (about both coefficients and states) on the corresponding factor graph.
A recently proposed ``turbo'' messaging schedule \cite{Schniter:CISS:10}
suggests to iterate between exploitation of HMT structure and exploitation of
observation structure from \eqref{system}.
For the former we use the standard sum-product algorithm \cite{Pearl:Book:88,Kschischang:TIT:01}, 
and for the latter we use the recently proposed  
\emph{approximate message passing} (AMP) approach \cite{Donoho:PNAS:09}.
The remarkable properties of AMP are 1) a rigorous analysis 
(as $M,N\rightarrow\infty$ with $M/N$ fixed, under i.i.d Gaussian $\vec{\Phi}$)
\cite{Bayati:TIT:11} establishing that its solutions are governed by a state-evolution whose fixed points---when unique---yield the true posterior means,
and 2) very low implementational complexity 
(e.g., AMP requires one forward and one inverse fast-wavelet-transform per 
iteration, and very few iterations).

We consider two types of conditional-Gaussian coefficient models:
a Bernoulli-Gaussian (BG) model and a two-state Gaussian-mixture (GM) model.
The BG model assumes that the coefficients are either generated from a 
large-variance Gaussian distribution or are exactly zero
(i.e., the coefficients are exactly sparse), whereas
the GM model assumes that the coefficients are generated from either 
a large-variance or a small-variance Gaussian distribution. 
Both models have been previously applied for imaging, e.g., 
the BG model was used in~\cite{HeCarin:SPL:10, HeCarin:TSP:09}, whereas
the GM model was used in~\cite{Crouse:TSP:98, Duarte:ICASSP:08}. 

Although our models for the coefficients $\vec{\theta}$ and the corresponding 
HMT states involve statistical parameters like variance and transition 
probability, we learn those parameters directly from the data.
To do so, we take a hierarchical Bayesian approach---similar to 
\cite{HeCarin:SPL:10, HeCarin:TSP:09}---where these statistical parameters are
treated as random variables with suitable hyperpriors. 
Experiments on a large image database show that our turbo-AMP approach yields 
state-of-the-art reconstruction performance with substantial reduction in complexity.

The remainder of the paper is organized as follows.
Section~\ref{sec:model} describes the signal model, 
Section~\ref{sec:alg} describes the proposed algorithm, 
Section~\ref{sec:sims} gives numerical results and comparisons with other algorithms,
and Section~\ref{sec:conc} concludes.

\emph{Notation}:
 Above and in the sequel, we use 
 lowercase boldface quantities to denote vectors, 
 uppercase boldface quantities to denote matrices, 
 $\vec{I}$ to denote the identity matrix,
 $(\cdot)^T$ to denote transpose,
 %
 %
 and 
 $\norm{\vec{x}}_2 \defn \sqrt{\vec{x}^T\vec{x}}$. 
 We use
 $p_{\Theta|S}(\theta\giv s)$ to denote the probability density\footnote{
   or the probability mass function (pmf), as will be clear from the context.} 
 function (pdf) of random variable $\Theta$ given the event $S=s$,
 where often the subscript ``$\mbox{}_{\Theta|S}$'' is omitted when there 
 is no danger of confusion.
 We use $\mc{N}(\vec{x};\vec{m},\vec{\Sigma})$
 to denote the $N$-dimensional Gaussian pdf with argument $\vec{x}$,
 mean $\vec{m}$, and covariance matrix $\vec{\Sigma}$,
 and we write $\vec{x}\sim\mc{N}(\vec{m},\vec{\Sigma})$ to indicate that random 
 vector $\vec{x}$ has this pdf.
 We use
 $\E\{\cdot\}$ to denote expectation, 
 $\Pr\{\mc{E}\}$ to denote the probability of event $\mc{E}$,
 %
 and $\delta(\cdot)$ to denote the Dirac delta.
 Finally, we use $\propto$ to denote equality up to a multiplicative constant.

\section{Signal Model}              \label{sec:model}
\begin{figure}[t]
\begin{center}
\begin{minipage}[b]{0.32\linewidth}
\centering
\includegraphics[width=\columnwidth]{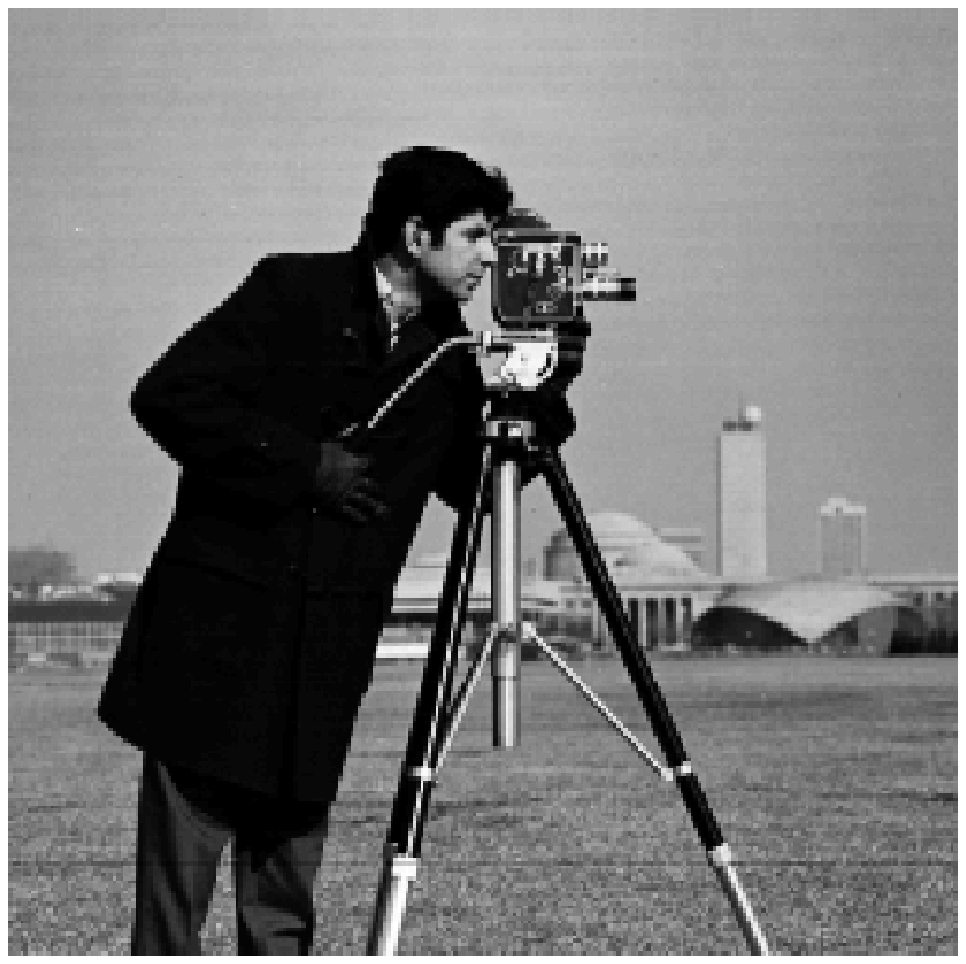}
\end{minipage}
\begin{minipage}[b]{0.32\linewidth}
\centering
\includegraphics[width=\columnwidth]{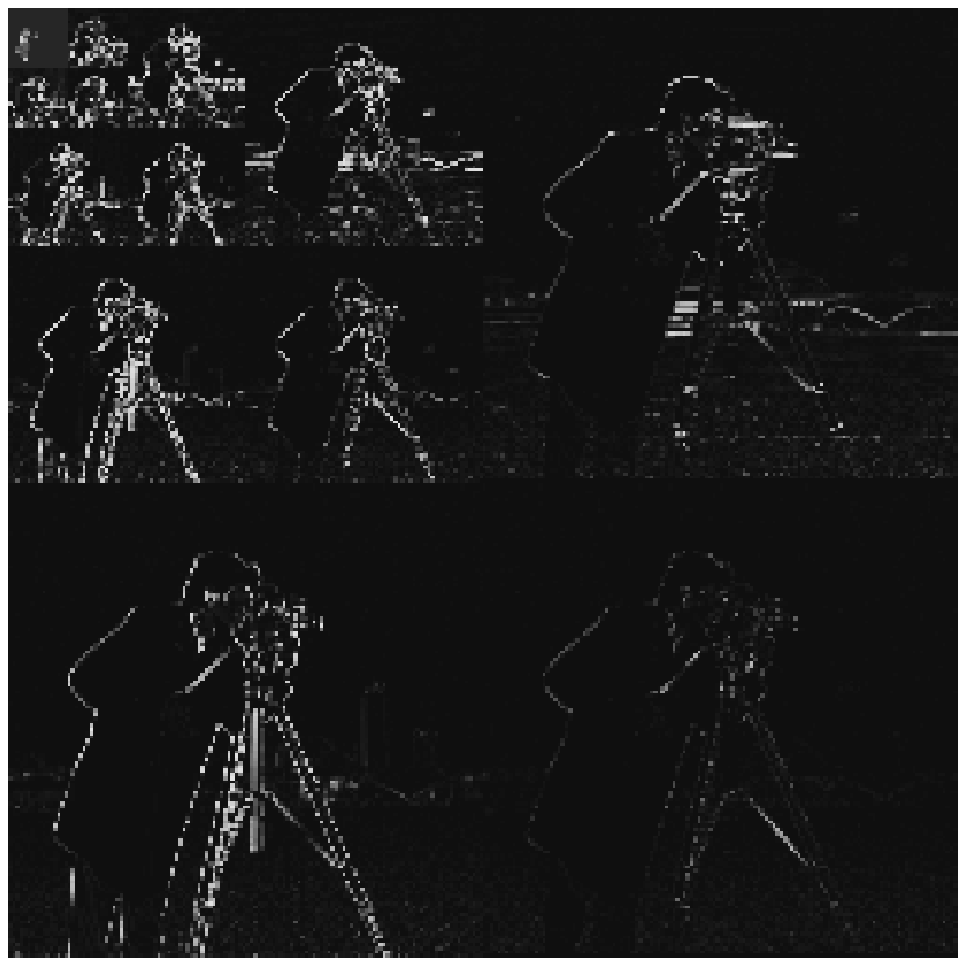}
\end{minipage}
\begin{minipage}[b]{0.32\linewidth}
\centering
\includegraphics[width=\columnwidth]{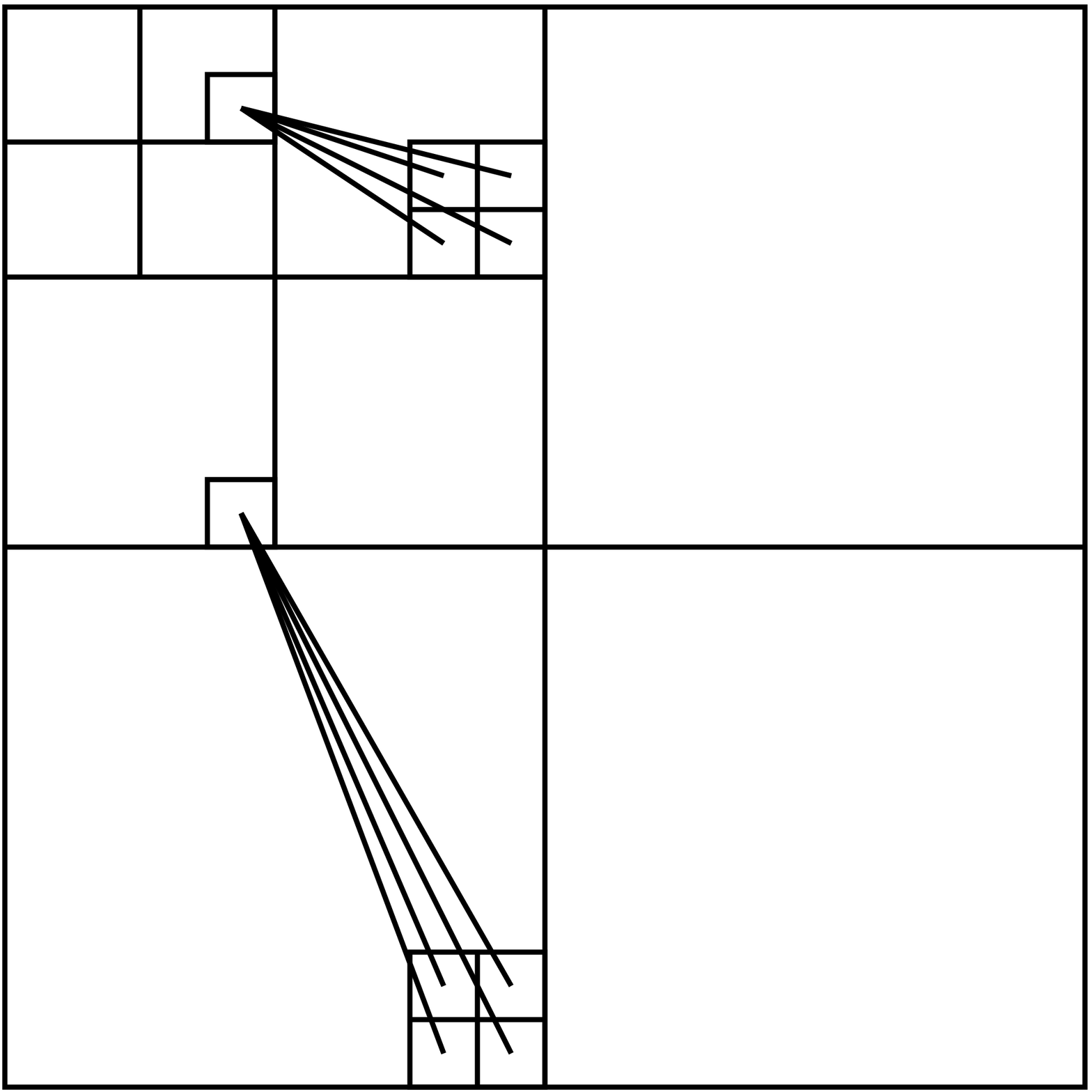}
\end{minipage}
\caption{Left: The camera-man image. Center: The corresponding transform 
coefficients, demonstrating PAS.  Right: An illustration of quad-tree structure.}
\label{fig:quadtree}
\end{center}
\end{figure}
Throughout, we assume that $\vec{\Psi}$ represents a 2D wavelet transform 
\cite{Mallat:Book:08}, 
so that the transform coefficients $\vec{\theta}=[\theta_1,\dots,\theta_N]^T$ 
can be 
partitioned into so-called ``wavelet'' coefficients (at indices $n\in\mc{W}$)
and ``approximation'' coefficients (at indices $n\in\mc{A}$).
The wavelet coefficients can be further partitioned into several quad-trees, each with 
$J\geq 1$ levels (see Fig.~\ref{fig:quadtree}). 
We denote the indices of all coefficients at level $j\in\{0,\dots,J\!-\!1\}$ 
of these wavelet trees by $\mc{W}_j$, where $j=0$ refers to the root.
In the interest of brevity, and with a slight abuse of notation, 
we refer to the approximation coefficients as level ``$-1$'' of the 
wavelet tree (i.e., $\mc{A}=\mc{W}_{-1}$).

\begin{figure}
\begin{center}
   \newcommand{\sz}{0.9}
   \psfrag{s1}[l][Bl][\sz]{$s_1$}
   \psfrag{s2}[l][Bl][\sz]{$s_2$}
   \psfrag{s3}[l][Bl][\sz]{$s_3$}
   \psfrag{s4}[l][Bl][\sz]{$s_4$}
   \psfrag{s5}[l][Bl][\sz]{$s_5$}
   \psfrag{s6}[l][Bl][\sz]{$s_6$}
   \psfrag{s7}[l][Bl][\sz]{$s_7$}
   \psfrag{s8}[l][Bl][\sz]{$s_8$}
   \psfrag{x1}[l][Bl][\sz]{$\theta_1$}
   \psfrag{x2}[l][Bl][\sz]{$\theta_2$}
   \psfrag{x3}[l][Bl][\sz]{$\theta_3$}
   \psfrag{x4}[l][Bl][\sz]{$\theta_4$}
   \psfrag{x5}[l][Bl][\sz]{$\theta_5$}
   \psfrag{x6}[l][Bl][\sz]{$\theta_6$}
   \psfrag{x7}[l][Bl][\sz]{$\theta_7$}
   \psfrag{x8}[l][Bl][\sz]{$\theta_8$}
   \psfrag{f1}[l][Bl][\sz]{$f_1$}
   \psfrag{f2}[l][Bl][\sz]{$f_2$}
   \psfrag{f3}[l][Bl][\sz]{$f_3$}
   \psfrag{f4}[l][Bl][\sz]{$f_4$}
   \psfrag{f5}[l][Bl][\sz]{$f_5$}
   \psfrag{f6}[l][Bl][\sz]{$f_6$}
   \psfrag{f7}[l][Bl][\sz]{$f_7$}
   \psfrag{f8}[l][Bl][\sz]{$f_8$}
   \psfrag{g1}[r][B][\sz]{$g_1$}
   \psfrag{g2}[r][B][\sz]{$g_2$}
   \psfrag{g3}[r][B][\sz]{$g_3$}
   \psfrag{p1}[l][Bl][\sz]{$q_1$}
   \psfrag{p5}[l][Bl][\sz]{$q_5$}
   \psfrag{p6}[l][Bl][\sz]{$q_6$}
   \psfrag{v}[l][Bl][\sz]{$\vdots$}
   \psfrag{SPE}[B][B][0.8]{\sf observation structure}
   \psfrag{SPD}[B][B][0.8]{\sf support structure}
    \epsfig{file=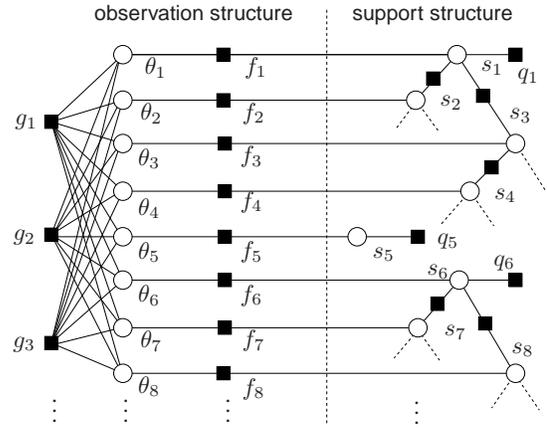,width=2.7in,clip=} 
\end{center}
\caption{Factor graph representation of the signal model. 
The variables $s_1$ and $s_6$ are wavelet states at the roots of two different 
Markov trees.
The variable $s_5$ is an approximation state and hence is not part of any Markov tree. 
The remaining $s_n$ are wavelet states at levels $j>0$. 
For visual simplicity, a binary-tree is shown instead of a quad-tree,
and the nodes representing the statistical parameters 
$\rho,\pi_{-1}^1,\pi_0^1,\{\rho_j,\pi_j^{11},\pi_j^{00}\}$, as well as those 
representing their hyperpriors, are not shown.}
\label{fig:FGA1}
\end{figure}

As discussed earlier, two coefficient models are considered in this paper: 
Bernoulli-Gaussian (BG) and two-state Gaussian mixture (GM).
For ease of exposition, we focus on the BG model until \secref{GM}, at which point
the GM case is detailed.
In the BG model, each transform coefficient $\theta_n$ is modeled using the 
(conditionally independent) prior pdf
\begin{eqnarray}
p(\theta_n\giv s_n) 
&=& s_n \mathcal{N}(\theta_n; 0, \sigma_n^2) + (1-s_n) \delta(\theta_n), \label{eqn:spike_slab}
\end{eqnarray}
where $s_n\in\{0,1\}$ is a hidden binary state.
The approximation states $\{s_n\}_{n\in\mc{W}_{-1}}$ are assigned the apriori activity rate 
$\pi_{-1}^1\defn \Pr\{s_n\!=\!1 \giv n\in\mc{W}_{-1}\}$, which is discussed further below.
Meanwhile, the root wavelet states $\{s_n\}_{n\in\mc{W}_0}$ are assigned 
$\pi_0^1\defn \Pr\{s_n\!=\!1 \giv n\in\mc{W}_0\}$.
Within each quad-tree, the states have a Markov structure. 
In particular, the activity of a state at level $j+1$ is determined by its 
parent's activity (at level $j$) and the transition probabilities
$\{\pi_j^{00},\pi_j^{11}\}$,
where $\pi_j^{00}$ denotes the probability that the child's state equals $0$ given that
his parent's state also equals $0$, and $\pi_j^{11}$ denotes the probability that the 
child's state equals $1$ given given that his parent's state also equals $1$.
The corresponding factor graph is shown in Fig.~\ref{fig:FGA1}.

We take a hierarchical Bayesian approach, modeling the statistical parameters 
$\sigma^2,\{\sigma_n^2\}_{n=1}^N,\pi_{-1}^1,\pi_{0}^1,\{\pi_j^{11},\pi_j^{00}\}_{j=0}^{J-2}$ 
as random variables and assigning them appropriate hyperpriors.
Rather than working directly with variances, we find it more convenient to work with precisions
(i.e., inverse-variances) such as $\rho\defn\sigma^{-2}$.
We then assume that all coefficients at the same level have the same precision, 
so that $\rho_j=\sigma_n^{-2}$ for all $n\in\mc{W}_{j}$.
To these precisions, we assign conjugate priors \cite{Gelman:Book:03}, which 
in this case take the form
\begin{eqnarray}
\rho &\sim& \mathrm{Gamma}(a,b) \\
\rho_j &\sim& \mathrm{Gamma}(a_j,b_j) 	\label{eqn:prior_rhoj},
\end{eqnarray}
where
$\mathrm{Gamma}(\rho;a,b) \defn \frac{1}{\Gamma(a)} b^{a} \rho^{a-1} \exp(-b \rho)$ for $\rho\geq 0$,
and where $a,b,\{a_j,b_j\}_{j=-1}^{J-1}$ are hyperparameters.
(Recall that the mean and variance of $\mathrm{Gamma}(a,b)$ are 
given by $a/b$ and $a/b^2$, respectively \cite{Gelman:Book:03}.)
For the activity rates and transition parameters, we assume
\begin{eqnarray}
\pi_0^1 &\sim& \mathrm{Beta}(c,d) \\
\pi_{-1}^1 &\sim& \mathrm{Beta}(\underline{c},\underline{d}) \\
\pi_j^{11} &\sim& \mathrm{Beta}(c_j,d_j) 	\label{eqn:prior_pij11}\\
\pi_j^{00} &\sim& \mathrm{Beta}(\underline{c}_j,\underline{d}_j) ,
\end{eqnarray}
where
$\mathrm{Beta}(p;c,d) 
 \defn \frac{\Gamma(c+d)}{\Gamma(c) \Gamma(d)} p^{c-1} (1-p)^{d-1}$, 
and where $c,d,\underline{c},\underline{d},
\{c_j,d_j,\underline{c}_j,\underline{d}_j\}_{j=1}^{J-1}$ are hyperparameters.
(Recall that the mean and variance of $\mathrm{Beta}(c,d)$ are given
by $\frac{c}{c+d}$ and $\frac{cd}{(c+d)^2 (c+d+1)}$, respectively 
\cite{Gelman:Book:03}.)
Our hyperparameter choices are detailed in \secref{sims}.

\section{Image Reconstruction}   \label{sec:alg}
To infer the wavelet coefficients $\vec{\theta}$, we would ideally like to 
compute the posterior pdf
\begin{eqnarray}
p(\ve{\theta}\giv \ve{y}) 
&\propto& \sum_{\vec{s}} p(\ve{y}\giv\ve{\theta},\ve{s}) p(\ve{\theta},\ve{s}) 
	\label{eqn:post1} \\[-3mm]
&=& \sum_{\vec{s}}\underbrace{p(\ve{s})}_{\displaystyle \triangleq h(\ve{s})} 
\prod_{n=1}^N \underbrace{p(\theta_n \giv s_n)}_{\displaystyle \triangleq f_n(\theta_n,s_n)} 
\prod_{m=1}^M \underbrace{p(y_m \giv \ve{\theta})}_{\displaystyle \triangleq g_m(\ve{\theta})}, 
	\quad \label{eqn:post2}
\end{eqnarray}
where $\propto$ denotes equality up to a multiplicative constant.
For the BG coefficient model, $f_n(\theta_n,s_n)$ is specified by \eqref{spike_slab}.
Due to the white Gaussian noise model \eqref{system}, we have
$g_m(\ve{\theta})=\sp{N}(y_m;\ve{a}_m^T\ve{\theta},\sigma^2)$,
where $\ve{a}_m^T$ denotes the $m^{\mathrm{th}}$ row of the matrix 
$\ve{A}\defn\ve{\Phi \Psi}$.

\subsection{Loopy Belief Propagation}			\label{sec:LBP}
While exact computation of $p(\vec{\theta}\giv\vec{y})$ is computationally
prohibitive, the marginal posteriors $\{p(\theta_n\giv\vec{y})\}$ can be 
efficiently approximated using \emph{loopy belief propagation} (LBP) 
\cite{Frey:ANIPS:98} on the factor graph of \figref{FGA1}, which 
uses round nodes to denote variables and square nodes to denote the
factors in \eqref{post2}.
In doing so, we also obtain the marginal posteriors $\{p(s_n\giv\vec{y})\}$.
For now, we treat statistical parameters 
$\rho,\pi_{-1}^1,\pi_0^1,\{\rho_j,\pi_j^{11},\pi_j^{00}\}$,
as if they were fixed and known,
and we detail the procedure by which they are learned in \secref{update}.

In LBP, messages are exchanged between the nodes of the factor graph until 
they converge.
Messages take the form of pdfs (or pmfs), 
and the message flowing to/from a variable node can be interpreted 
as a local belief about that variable.
According to the \emph{sum-product algorithm} \cite{Pearl:Book:88,Kschischang:TIT:01}
the message emitted by a variable node along a given edge is (an appropriate 
scaling of) the product of the incoming messages on all other edges. 
Meanwhile, the message emitted by a function node along a given edge is (an 
appropriate scaling of) the integral (or sum) of the product of the node's 
constraint function and the incoming messages on all other edges, where
the integration (or summation) is performed over all variables other than the one
directly connected to the edge along which the message travels. 
When the factor graph has no loops, exact marginal posteriors result from
two (i.e., forward and backward) passes of the sum-product algorithm
\cite{Pearl:Book:88,Kschischang:TIT:01}.
When the factor graph has loops, however, exact inference is known to be 
NP hard \cite{Cooper:AI:90} and so LBP is not guaranteed to produce correct 
posteriors.
Still, LBP has shown state-of-the-art performance in many applications,
such as 
inference on Markov random fields \cite{Freeman:IJCV:00},
turbo decoding \cite{McEliece:JSAC:98},
LDPC decoding \cite{MacKay:Book:03},
multiuser detection \cite{Boutros:TIT:02}, 
and compressive sensing \cite{Baron:TSP:10,Donoho:PNAS:09,Donoho:ITW:10a,Bayati:TIT:11}.

\subsection{Message Scheduling: The Turbo Approach}\label{sec:MP:Turbo}
\begin{figure}[t]
\begin{center}
\newcommand{\sz}{0.8}
\psfrag{p}[l][Bl][\sz]{$h$}
\psfrag{g1}[l][Bl][\sz]{$g_1$}
\psfrag{g2}[l][Bl][\sz]{$g_2$}
\psfrag{gM}[l][Bl][\sz]{\hspace{-1mm}$g_M$}
\psfrag{x1}[l][Bl][\sz]{$\theta_1$}
\psfrag{x2}[l][Bl][\sz]{$\theta_2$}
\psfrag{x3}[l][Bl][\sz]{$\theta_3$}
\psfrag{xN}[l][Bl][\sz]{$\theta_N$}
\psfrag{q1}[l][Bl][\sz]{$f_1$}
\psfrag{q2}[l][Bl][\sz]{$f_2$}
\psfrag{q3}[l][Bl][\sz]{$f_3$}
\psfrag{qN}[l][Bl][\sz]{$f_N$}
\psfrag{s1}[l][Bl][\sz]{$s_1$}
\psfrag{s2}[l][Bl][\sz]{$s_2$}
\psfrag{s3}[l][Bl][\sz]{$s_3$}
\psfrag{sN}[l][Bl][\sz]{$s_N$}
\psfrag{r1}[l][Bl][\sz]{$d_1\of{t}$}
\psfrag{r2}[l][Bl][\sz]{$d_2\of{t}$}
\psfrag{r3}[l][Bl][\sz]{$d_3\of{t}$}
\psfrag{rN}[l][Bl][\sz]{$d_N\of{t}$}
\psfrag{h1}[l][Bl][\sz]{$h_1\of{t}$}
\psfrag{h2}[l][Bl][\sz]{$h_2\of{t}$}
\psfrag{h3}[l][Bl][\sz]{$h_3\of{t}$}
\psfrag{hN}[l][Bl][\sz]{$h_N\of{t}$}
\psfrag{l}[][B][1.0]{$\vdots$~}
\psfrag{AMP}[B][B][0.8]{\sf AMP}
\includegraphics[width=0.8\columnwidth]{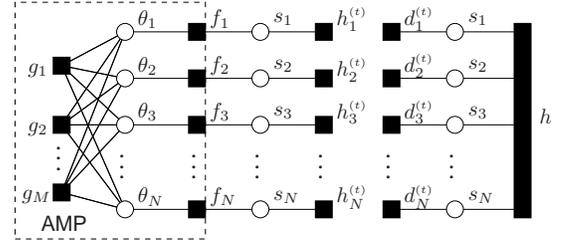}
\end{center}
\caption{The turbo approach yields a decoupled factor graph.}
\label{fig:Turbo}
\end{figure}

With loopy belief propagation, there exists some freedom in
how messages are scheduled.
In this work, we adopt the ``turbo'' approach recently proposed 
in \cite{Schniter:CISS:10}.
For this, we split the factor graph in Fig.~\ref{fig:FGA1} along the dashed line
and obtain the two decoupled subgraphs in Fig.~\ref{fig:Turbo}. 
We then alternate between belief propagation on each of these two subgraphs,
treating the likelihoods on $\{s_n\}$ generated from belief propagation on
one subgraph as priors for subsequent belief propagation on the other 
subgraph.
We now give a more precise description of this turbo scheme, referring to one 
full round of alternation as a ``turbo iteration.''
In the sequel, we use $\nu_{A \to B}\of{t}(.)$ to denote the message passed 
from node $A$ to node $B$ during the $t^\mathrm{th}$ turbo iteration. 

The procedure starts at $t=1$ by setting the ``prior'' pmfs 
$\{h_n\of{1}(.)\}$ in accordance with the apriori activity rates 
$\Pr\{s_n=1\}$ described in \secref{model}.
LBP is then iterated (to convergence) on the left subgraph in \figref{Turbo},
finally yielding the messages $\{\nu_{f_n\to s_n}\of{1}(.)\}$. 
We note that the message $\nu_{f_n\to s_n}\of{1}(s_n)$ can be interpreted 
as the current estimate of the likelihood\footnote{
  In turbo decoding parlance, the likelihood
  $\nu_{f_n\to s_n}\of{t}(s_n)$ would be referred to as the ``extrinsic'' 
  information about $s_n$ produced by the left ``decoder'', since it 
  does not directly involve the corresponding prior $h_n\of{t}(s_n)$.
  Similarly, the message $\nu_{h \to s_n}^{(t)}(s_n)$ would be referred
  to as the extrinsic information about $s_n$ produced by the right decoder.}
on $s_n$, i.e., $p(\vec{y}\giv s_n)$ as a function of $s_n$.
These likelihoods are then treated as priors for belief 
propagation on the right subgraph, as facilitated by the assignment 
$d_n\of{1}(.) = \nu_{f_n \to s_n}\of{1}(.)$ for each $n$.
Due to the tree structure of HMT, there are no loops in right subgraph
(i.e., inside the ``$h$'' super-node in \figref{Turbo}), 
and thus it suffices to perform only one forward-backward pass of the
sum-product algorithm \cite{Pearl:Book:88,Kschischang:TIT:01}.
The resulting leftward messages $\nu_{h \to s_n}^{(1)}(.)$ are 
subsequently treated as priors for belief propagation on the left 
subgraph at the next turbo iteration, as facilitated by the assignment 
$h_n^{(2)}(.) = \nu_{h \to s_n}^{(1)}(.)$.
The process then continues for turbo iterations $t=2,3,4,\dots$, until
the likelihoods converge or a maximum number of turbo iterations has
elapsed.
Formally, the turbo schedule is summarized by
\begin{eqnarray}
 d_n^{(t)}(s_n) &=& \nu_{f_n \to s_n}^{(t)}(s_n) \\
 h_n^{(t+1)}(s_n) &=& \nu_{h \to s_n}^{(t)}(s_n) .
\end{eqnarray}

In the sequel, we refer to inference of $\{s_n\}$ using 
compressive-measurement structure (i.e., inference on the left subgraph
of \figref{Turbo}) as 
{\em soft support-recovery} (SSR) and inference of $\{s_n\}$ using 
HMT structure (i.e., inference on the right subgraph of \figref{Turbo}) as 
{\em soft support-decoding} (SSD). 
SSR details are described in the next subsection.

\subsection{Soft Support-Recovery via AMP}\label{sec:MP:SSR}
We now discuss our implementation of SSR during a single turbo iteration $t$. 
Because the operations are invariant to $t$, we suppress the $t$-notation.
As described above, SSR performs several iterations of loopy belief propagation
per turbo iteration using the fixed priors $\lambda_n\defn h_n(s_n=1)$.
This implies that, over SSR's LBP iterations, the message 
$\nu_{f_n\to\theta_n}(.)$ is fixed at 
\begin{eqnarray}
\nu_{f_n \to \theta_n}(\theta_n) 
&=& \lambda_n \sp{N}(\theta_n;0,\sigma_n^2) + (1-\lambda_n)\delta(\theta_n)  .\label{eqn:fnToxn}
\end{eqnarray}
The dashed box in \figref{Turbo} shows the region of the factor graph on which 
messages are updated during SSR's LBP iterations.
This subgraph can be recognized as the one that Donoho, Maleki, and Montanari 
used to derive their so-called \emph{approximate message passing} (AMP) 
algorithm \cite{Donoho:PNAS:09}.
While \cite{Donoho:PNAS:09} assumed an i.i.d Laplacian prior for 
$\vec{\theta}$, the approach for generic i.i.d priors was outlined 
in \cite{Donoho:ITW:10a}.
Below, we extend the approach of \cite{Donoho:ITW:10a} to independent
\emph{non}-identical priors (as analyzed in \cite{Som:NAECON:10})
and we detail the Bernoulli-Gaussian case.
In the sequel, we use a superscript-$i$ to index SSR's LBP iterations.

According to the sum-product algorithm, the fact that $\nu_{f_n\to\theta_n}(.)$ 
is non-Gaussian implies that $\nu^i_{\theta_n \to g_m}(.)$ is also non-Gaussian,
which complicates the exact calculation of the subsequent messages 
$\nu^i_{g_m \to \theta_n}(.)$ as defined by the sum-product algorithm.
However, for large $N$, the combined effect of 
$\{\nu^i_{\theta_n \to g_m}(.)\}_{n=1}^N$ at the $g_m$ nodes
can be approximated as Gaussian
using central-limit theorem (CLT) arguments, after which it becomes 
sufficient to parameterize each message $\nu^i_{\theta_n \to g_m}(.)$ 
by only its mean and variance:
\begin{eqnarray}
\mu_{mn}^i 
&\triangleq& \textstyle \int_{\theta_n} \theta_n\, \nu^i_{\theta_n \to g_m}(\theta_n) 
	\label{eqn:mu_mn} \\
v_{mn}^i 
&\triangleq& \textstyle \int_{\theta_n} (\theta_n-\mu^i_{mn})^2 \,\nu^i_{\theta_n \to g_m}(\theta_n). 
	\label{eqn:v_mn}
\end{eqnarray}
Combining 
\begin{eqnarray}
\prod_q \sp{N}(\theta;\mu_q,v_q) 
&\propto& \sp{N}\left(\theta;\frac{\sum_q \mu_q/v_q}{\sum_q 1/v_q},
	\frac{1}{\sum_q 1/v_q}\right) 	\label{eqn:prod_Gauss}
\end{eqnarray}
with $g_m(\ve{\theta})=\sp{N}(y_m;\ve{a}^T_m \ve{\theta},\sigma^2)$, 
the CLT then implies that
\begin{eqnarray}
\nu^i_{g_m \to \theta_n}(\theta_n) 
&\approx& \sp{N} \left(\theta_n; \frac{z_{mn}^i}{A_{mn}}, \frac{c^i_{mn}}{A_{mn}^2} \right) \label{eqn:gmToxn} \\
z^i_{mn} &\triangleq& \textstyle y_m - \sum_{q \neq n} A_{mq} \mu^i_{qm} \label{eqn:z_mn} \\
c^i_{mn} &\triangleq& \textstyle \sigma^2 + \sum_{q \neq n} A_{mq}^2 v^i_{qm}. \label{eqn:c_mn}
\end{eqnarray}
The updates $\mu^{i+1}_{mn}$ and $v^{i+1}_{mn}$ can then be calculated from 
\begin{eqnarray}
\nu^{i+1}_{\theta_n \to g_m}(\theta_n) 
&\propto& \textstyle \nu_{f_n \to \theta_n}(\theta_n) \prod_{l \neq m} 
	\nu^i_{g_l \to \theta_n}(\theta_n), \label{eqn:xnTogm_ip1}
\end{eqnarray}
where, using \eqref{prod_Gauss}, the product term in \eqref{xnTogm_ip1} is 
\begin{eqnarray}
\propto \sp{N}\left(\theta_n; \frac{\sum_{l \neq m} A_{ln}z^i_{ln}/c^i_{ln}}{\sum_{l \neq m} A^2_{ln}z^i_{ln}/c^i_{ln}},\frac{1}{\sum_{l \neq m}A^2_{ln}/c^i_{ln}} \right). \label{eqn:prod1}
\end{eqnarray}
Assuming that the values $A_{ln}^2$ satisfy
\begin{eqnarray}
\sum_{l \neq m}A^2_{ln} \approx \sum_{l=1}^M A^2_{ln} \approx 1, 
\end{eqnarray}
which occurs, e.g., when $M$ is large and $\{A_{ln}\}$ are generated 
i.i.d with variance $1/M$, we have
$c^i_{ln} \approx \textstyle c^i_n \triangleq \frac{1}{M} \sum_{m=1}^M c^i_{mn}$, and thus \eqref{xnTogm_ip1} is well approximated by 
\begin{eqnarray}
\nu^{i+1}_{\theta_n \to g_m}(\theta_n) 
&\propto& \left(\lambda_n \sp{N}(\theta_n;0,\sigma_n^2) + (1-\lambda_n)\delta(\theta_n)\right) 
\nonumber\\ && \mbox{}\times 
\sp{N}(\theta_n; \xi_{nm}^i, c_n^i) \label{eqn:xnTogm_ip1_appx} \\
\xi^i_{nm} 
&\triangleq& \textstyle \sum_{l \neq m} A_{ln} z^i_{ln}. \label{eqn:xi_mn}
\end{eqnarray}
In this case, the mean and variance of 
$\nu^{i+1}_{\theta_n \to g_m}(.)$ become
\begin{eqnarray}
\mu_{nm}^{i+1} &=& \alpha_n(c_n^i) \xi_{nm}^i / (1+\gamma_{nm}^i) \\
v_{nm}^{i+1} &=& \gamma_{nm}^i (\mu_{nm}^{i+1})^2 + \mu_{nm}^{i+1} c_n^i / \xi_{nm}^i \\
\gamma_{nm}^i &\triangleq& \beta_n(c_n^i) \exp(-\zeta_n(c_n^i)(\xi_{nm}^i)^2), \\[-8mm]\nonumber
\end{eqnarray}
where 
\begin{eqnarray*}
\alpha_n(c) 
&\defn& (c/\sigma_n^2 + 1)^{-1} \\
\beta_n(c) 
&\defn& (-1+1/\lambda_n) \sqrt{1+\sigma_n^2/c} \\ 
\zeta_n(c) 
&\defn& (2c(1+c/\sigma_n^2))^{-1}. 
\end{eqnarray*}
According to the sum-product algorithm, $\hat{p}^{i+1}(\theta_n\giv\ve{y})$,
the posterior on $\theta_n$ after SSR's $i^{\mathrm{th}}$-LBP iteration, obeys
\begin{eqnarray}
\hat{p}^{i+1}(\theta_n\giv\ve{y}) 
&\propto& \textstyle \nu_{f_n \to \theta_n}(\theta_n) 
	\prod_{l=1}^M \nu^i_{g_l \to \theta_n}(\theta_n), \label{eqn:p_hat_xn}
\end{eqnarray}
whose mean and variance determine the $i^{\mathrm{th}}$-iteration MMSE estimate of $\theta_n$ 
and its variance, respectively. 
Noting that the difference between~\eqref{p_hat_xn} and \eqref{xnTogm_ip1} is only the 
inclusion of the $m^{\mathrm{th}}$ product term, these MMSE quantities become
\begin{eqnarray}
\mu_{n}^{i+1} &=& \alpha_n(c_n^i) \xi_{n}^i / (1+\gamma_{n}^i) \\
v_{n}^{i+1} &=& \gamma_{n}^i (\mu_{n}^{i+1})^2 + \mu_{n}^{i+1} c_n^i / \xi_{n}^i\\
\xi^i_{n} &\triangleq& \textstyle \sum_{l=1}^M A_{ln} z^i_{ln}\\
\gamma_{n}^i &\triangleq& \beta_n(c_n^i) \exp(-\zeta_n(c_n^i)(\xi_{n}^i)^2).
\end{eqnarray}
Similarly, the posterior on $s_n$ after the $i^{\mathrm{th}}$ iteration obeys
\begin{eqnarray}
\hat{p}^{i+1}(s_n \giv \ve{y})
&\propto& \nu^{i}_{f_n \to s_n}(s_n) \nu_{h_n \to s_n}(s_n), 
\\[-7mm]\nonumber
\end{eqnarray}
where
\vspace{-5mm}
\begin{eqnarray}
\nu^i_{f_n \to s_n}(s_n)
&\propto& \int_{\theta_n} f_n(\theta_n,s_n) \prod_{l=1}^M \nu^i_{g_l \to \theta_n}(\theta_n).
\\[-4mm]\nonumber
\end{eqnarray}
Since $f_n(\theta_n,s_n)=s_n \sp{N}(\theta_n;0,\sigma_n^2) + (1-s_n)\delta(\theta_n)$, 
it can be seen that the corresponding log-likelihood ratio (LLR) is 
\begin{eqnarray}
L^{i+1}_n 
\triangleq \ln \frac{\nu_{f_n \to s_n}^i(s_n\!=\!1)}{\nu_{f_n \to s_n}^i(s_n\!=\!0)} 
= \ln \frac{1-\lambda_n^i}{\gamma_n^i \lambda_n^i}. \quad \label{eqn:amp_llr}
\end{eqnarray}
Clearly, the LLR $L_n^{i+1}$ and the likelihood function 
$\nu_{f_n \to s_n}^i(.)$ express the same information, but in different ways.

The procedure described thus far updates $\sp{O}(MN)$ variables
per LBP iteration, which is impractical since $MN$ can be very large.
In \cite{Donoho:ITW:10a}, Donoho, Maleki, and Montanari proposed, for the
i.i.d case, further approximations that yield a ``first-order'' approximate 
message passing (AMP) algorithm that allows the update of only $\sp{O}(N)$ 
variables per LBP iteration, essentially by approximating the 
\emph{differences} among the outgoing means/variances of the $g_m(\cdot)$ 
nodes (i.e., $z^i_{mn}$ and $c^i_{mn}$) as well as the differences among 
the outgoing means/variances of the $\theta_{n}$ nodes (i.e., 
$\mu_n^{i}$ and $v_n^i$).
These resulting algorithm was then rigorously analyzed by Bayati
and Montanari in \cite{Bayati:TIT:11}.
We now summarize a straightforward extension of the i.i.d AMP algorithm from
\cite{Donoho:ITW:10a} to the case of an independent but non-identical 
Bernoulli-Gaussian prior \eqref{fnToxn}: 
\begin{eqnarray}
\xi^i_n &=& \textstyle \sum_{m=1}^M A_{mn}z^i_m + \mu^i_n \label{eqn:xini}\\
\mu^{i+1}_n &=& F_n(\xi^i_n;c^i) \\
v^{i+1}_n &=& G_n(\xi^i_n;c^i) \\
z^{i+1}_m &=& \textstyle y_m - \sum_{n=1}^N A_{mn} \mu^i_n + \frac{z^i_m}{M} \sum_{n=1}^N F'_n(\xi^i_n;c^i)\\
c^{i+1} &=& \textstyle \sigma^2 + \frac{1}{M} \sum_{n=1}^N v^{i+1}_n, \label{eqn:cni} 
\end{eqnarray}
where $F_n(.;.)$, $G_n(.;.)$ and $F'_n(.;.)$ are defined as
\begin{eqnarray}
F_n(\xi;c) &=& \frac{\alpha_n(c)}{1+\tau_n(\xi;c)} \xi \label{eqn:Fn}\\
G_n(\xi;c) &=& \tau_n(\xi;c) F_n(\xi;c)^2 + c\,\xi^{-1}\! F_n(\xi;c) \label{eqn:Gn}\\
F'_n(\xi;c) &=& \frac{\alpha_n(c)\big[1 + \tau_n(\xi;c) \left(1+2\zeta(c_n)\xi^2 \right)\!\!\big]}{[1+\tau_n(\xi;c)]^2} \! \quad \label{eqn:Fnd}\\
\tau_n(\xi;c) &=& \beta_n(c) \exp(-\zeta_n(c) \xi^2). \label{eqn:taun}
\end{eqnarray}

For the first turbo iteration (i.e., $t\!=\!1$), we initialize AMP using $z^{i=1}_{m}=y_m$, $\mu^{i=1}_n=0$, and $c^{i=1}_n \gg \sigma_n^2$ for all $m, n$. 
For subsequent turbo iterations (i.e., $t\!>\!1$), we initialize AMP by setting $z^{i=1}_{m}, \mu^{i=1}_n, c^{i=1}_n$ equal to the final values of $z^{i}_{m}, \mu^{i}_n, c^{i}_n$ generated by AMP at the \emph{previous} turbo iteration.
We terminate the AMP iterations as soon as either $\norm{\vec{\mu}^i-\vec{\mu}^{i-1}}_2<10^{-5}$ or a maximum of $10$ AMP iterations have elapsed.
Similarly, we terminate the turbo iterations as soon as either $\norm{\vec{\mu}\of{t}-\vec{\mu}\of{t-1}}_2<10^{-5}$ a maximum of $10$ turbo iterations have elapsed.
The final value of $\vec{\mu}=[\mu_1,\dots,\mu_N]^T$ is output at the signal estimate $\hvec{\theta}$.

\subsection{Learning the Statistical Parameters} \label{sec:update}
We now describe how the precisions $\{\rho_j\}$ are learned.
First, we recall that $\rho_j$ describes the apriori precision on the active
coefficients at the $j^{th}$ level, i.e., on $\{\theta_n\}_{n\in\mc{S}_j}$,
where the corresponding index set $\mc{S}_j\defn \{n\in\mc{W}_j: s_n=1\}$ 
is of size $K_j\defn |\mc{S}_j|$.
Furthermore, we recall that the prior on $\rho_j$ was chosen as 
in \eqref{prior_rhoj}.
Thus, \emph{if} we had access to the true values 
$\{\theta_n\}_{n\in\mc{S}_j}$, then \eqref{spike_slab} implies that
\begin{eqnarray}
  p(\theta_n\giv n\in\mc{S}_j) 
  &=& \mc{N}(\theta_n;0,\rho_j) ,			\label{eqn:theta_on}
\end{eqnarray}
which implies\footnote{
  This posterior results because the chosen prior is conjugate 
  \cite{Gelman:Book:03} for the likelihood in \eqref{theta_on}.} 
that the posterior on $\rho_j$ would take the form of 
$\mathrm{Gamma}(\hat{a}_j,\hat{b}_j)$ where
$\hat{a}_j=a_j+\frac{1}{2}K_j$ and $\hat{b}_j=b_j+
\frac{1}{2}\sum_{n\in\mc{S}_j}\theta_n^2$. 
In practice, we don't have access to the true values 
$\{\theta_n\}_{n\in\mc{S}_j}$ nor to the set $\mc{S}_j$, and thus we propose 
to build surrogates from the SSR outputs.
In particular, to update $\rho_j$ after the $t^\mathrm{th}$ turbo iteration, 
we employ
\begin{eqnarray}
\mc{S}\of{t}_j
&\defn& \{n\in\mc{W}_j: L\of{t}_n>0\} \\
K\of{t}_j
&\defn& |\mc{S}\of{t}_j| ,		\label{eqn:Kj}
\end{eqnarray}
and $\{\mu_n\of{t}\}_{n\in\mc{S}\of{t}_j}$, where 
$L\of{t}_n$ and $\mu\of{t}_n$
denote the final LLR on $s_n$ and the final MMSE estimate of $\theta_n$,
respectively, at the $t^\mathrm{th}$ turbo iteration.
These choices imply the hyperparameters 
\begin{eqnarray}
\hat{a}_j\of{t+1} &=& \textstyle a_j + \frac{1}{2}K_j\of{t} \\
\hat{b}_j\of{t+1} &=& \textstyle b_j + \frac{1}{2} \sum_{n\in\mc{S}\of{t}_j}(\mu_n\of{t})^2.
\end{eqnarray}
Finally, to perform SSR at turbo iteration $t+1$, we set the variances 
$\{\sigma_n^2\}_{n\in\mc{W}_j}$ equal to the inverse of the expected 
precisions, i.e., $1/\E\{\rho_j\}=\hat{b}_j\of{t+1}/\hat{a}_j\of{t+1}$.
The noise variance $\sigma^2$ is learned 
similarly from the SSR-estimated residual.

Next, we describe how the transition probabilities $\{\pi_j^{11}\}$ are 
learned.
First, we recall that $\pi_j^{11}$ describes the probability that a child
at level $j+1$ is active (i.e., $s_n=1$) given that his parent (at level $j$)
is active.
Furthermore, we recall that the prior on $\pi_j^{11}$ was chosen as
in \eqref{prior_pij11}.
Thus \emph{if} we knew that there were $K_j$ active coefficients at level $j$, 
of which $C_j$ had active children, then\footnote{
  This posterior results because the chosen prior is conjugate to the
  Bernoulli likelihood \cite{Gelman:Book:03}.} 
the posterior on $\pi_j^{11}$ 
would take the form of $\mathrm{Beta}(\hat{c}_j,\hat{d}_j)$, where 
$\hat{c}_j = c_j + C_j$ and 
$\hat{d}_j = d_j + K_j - C_j$.
In practice, we don't have access to the true values of $K_j$ and $C_j$,
and thus we build surrogates from the SSR outputs.
In particular, to update $\pi_j^{11}$ after the $t^\mathrm{th}$ turbo
iteration, we approximate $s_n=1$ by the event $L\of{t}_n>0$, and based on 
this approximation set $K\of{t}_j$ (as in \eqref{Kj}) and $C\of{t}_j$.
The corresponding hyperparameters are then updated as 
\begin{eqnarray}
\hat{c}_j\of{t+1} 
&=& c_j + C_j\of{t}\\
\hat{d}_j\of{t+1} 
&=& d_j + K_j\of{t} - C_j\of{t}.
\end{eqnarray}
Finally, to perform SSR at turbo iteration $t+1$, we set the transition
probabilities ${\pi}_j^{11}$ equal to the expected value
$\hat{c}_j\of{t+1}/( \hat{c}_j\of{t+1}+\hat{d}_j\of{t+1} )$.
The parameters $\pi_0^1$, $\pi_{-1}^1$, and $\{\pi_j^{00}\}$ are learned 
similarly.

\subsection{The Two-State Gaussian-Mixture Model}	\label{sec:GM}
Until now, we have focused on the Bernoulli-Gaussian (BG) signal model 
\eqref{spike_slab}.
In this section, we describe the modifications needed to handle the
Gaussian mixture (GM) model 
\begin{eqnarray}
p(\theta_n\giv s_n) 
&=& s_n \mathcal{N}(\theta_n; 0, \sigma\nL^2) + (1-s_n)\mathcal{N}(\theta_n; 0, \sigma\nS^2), \nonumber\\
\label{eqn:gmm}
\end{eqnarray}
where $\sigma\nL^2$ denotes the variance of ``large'' coefficients
and $\sigma\nS^2$ denotes the variance of ``small'' ones.
For either the BG or GM prior, AMP is performed using the steps 
\eqref{xini}-\eqref{cni}.
For the BG case, the functions $F_n(.;.)$, $G_n(.;.)$, $F'_n(.;.)$, 
and $\tau_n(.;.)$ are given in \eqref{Fn}--\eqref{taun}, whereas for the 
GM case, they take the form
\begin{eqnarray}
F_n(\xi;c) &=& \frac{\bar{\alpha}\nL(c)+\bar{\alpha}\nS(c)\bar{\tau}_n(\xi,c)}{1+\bar{\tau}_n(\xi,c)} \xi \label{eqn:Fn2}\\
G_n(\xi;c) &=& \frac{\bar{\tau}_n(\xi,c) \xi^2 [\bar{\alpha}\nL(c)-\bar{\alpha}\nS(c)]^2}{[1+\bar{\tau}_n(\xi,c)]^2}  +\, c\,\xi^{-1}\! F_n(\xi; c) \nonumber\label{eqn:Gn2}\\[-1.5mm]&&\\[-3.0mm]
F'_n(\xi;c) &=& \frac{\bar{\tau}_n(\xi;c)\bar{\alpha}\nL(c)(1+\bar{\tau}_n(\xi;c) - 2 \xi^2 \bar{\zeta}_n(c)) }{[1+\bar{\tau}_n(\xi;c)]^2} \nonumber\\ &&+  \frac{\bar{\alpha}\nS(c)(1+\bar{\tau}_n(\xi;c) + 2 \xi^2 \bar{\zeta}_n(c)\bar{\tau}_n(\xi;c)) }{[1+\bar{\tau}_n(\xi;c)]^2}  \quad \label{eqn:Fnd2}\\
\bar{\tau}_n(\xi,c) &=& \bar{\beta}_n(c) \exp(-\bar{\zeta}_n(c)\xi^2),
\end{eqnarray}
where
\begin{eqnarray}
\bar{\alpha}\nL(c) &\defn& \frac{\sigma\nL^2}{c + \sigma\nL^2}\\
\bar{\alpha}\nS(c) &\defn& \frac{\sigma\nS^2}{c + \sigma\nS^2}\\
\bar{\beta}_{n}(c) &\defn& \frac{1-\lambda_n}{\lambda_n} \sqrt{\frac{c+\sigma\nL^2}{c+\sigma\nS^2}}\\
\bar{\zeta}_{n}(c) &\defn& \frac{\sigma\nL^2 - \sigma\nS^2}{2(c+\sigma\nL^2)(c+\sigma\nS^2)}.
\end{eqnarray}
Likewise, for the BG case, the extrinsic LLR is given by \eqref{amp_llr}, 
whereas for the GM case, it becomes
\begin{eqnarray}
L^{i+1}_n 
\triangleq \ln \frac{\nu_{f_n \to s_n}^i(s_n\!=\!1)}{\nu_{f_n \to s_n}^i(s_n\!=\!0)} 
= \ln \frac{1-\lambda_n^i}{\bar{\tau}_n(\xi_n^i;c_n^i) \lambda_n^i}. \quad \label{eqn:amp_llr2}
\end{eqnarray}


\section{Numerical Results}				\label{sec:sims} 
\subsection{Setup}
The proposed turbo\footnote{An implementation of our algorithm can be downloaded from 
    {\sl \url{http://www.ece.osu.edu/~schniter/turboAMPimaging}}}
approach to compressive imaging was compared to several 
other tree-sparse reconstruction algorithms: 
ModelCS~\cite{Baraniuk:TIT:10}, 
HMT+IRWL1~\cite{Duarte:ICASSP:08},
MCMC~\cite{HeCarin:TSP:09},
variational Bayes~(VB)~\cite{HeCarin:SPL:10}; 
and to several simple-sparse reconstruction algorithms:
CoSaMP~\cite{Needell:ACHA:09}, 
SPGL1~\cite{vandenBerg:JSC:08}, and
Bernoulli-Gaussian (BG) AMP.
All numerical experiments were performed on $128\! \times\! 128$ (i.e., $N\!=\!16384$)
grayscale images using a $J=4$-level 2D Haar wavelet decomposition,
yielding $8^2\!=\!64$ approximation coefficients and 
$3\! \times\! 8^2\! =\! 192$ individual Markov trees.
In all cases, the measurement matrix $\vec{\Phi}$ had i.i.d Gaussian entries.
Unless otherwise specified, $M\!=\!5000$ noiseless measurements were used.
We used normalized mean squared error (NMSE) 
$\|\ve{x}-\ve{\hat{x}}\|_2^2 / \|\ve{x}\|_2^2$ as the performance metric.

We now describe how the hyperparameters were chosen for the proposed 
Turbo schemes.
Below, we use $N_j$ to denote the total number of wavelet coefficients 
at level $j$, and $N_{-1}$ to denote the total number of approximation 
coefficients.
For both Turbo-BG and Turbo-GM,
the Beta hyperparameters were chosen so that 
$c\!+\!d\!=\!N_0$, 
$\underline{c}\!+\!\underline{d}\!=\!N_{-1}$  
and $c_j\!+\!d_j\!=\!N_j~\forall j$ 
with
$\E\{p_0^1\}\!=\!1/N$, 
$\E\{p_{-1}^1\}\!=\!1-10^{-6}$, 
$\E\{p_j^{00}\}\!=\!1/N~\forall j$, and 
$\E\{p_j^{11}\}\!=\!0.5~\forall j$. 
These informative hyperparameters are similar to the ``universal'' 
recommendations in \cite{Romberg:TIP:01} and, in fact, identical to the 
ones suggested in the MCMC work \cite{HeCarin:TSP:09}.
%
For Turbo-BG, the hyperparameters for the signal precisions 
$\{\sigma^{-2}_n\}_{n\in\mc{W}_j}$ were set to $a_j\!=\!1~\forall j$ and 
$[b_{-1},\dots,b_3]\!=\![10, 1, 1, 0.1, 0.1]$.
This choice is motivated by the fact that wavelet coefficient magnitudes 
are known to decay exponentially with scale $j$ (e.g., \cite{Romberg:TIP:01}).
Meanwhile, the hyperparameters for the noise precision $\sigma^{-2}$ 
were set to $(a,b)=(1,10^{-6})$. 
Although the measurements were noiseless, we allow Turbo-BG a nonzero noise 
variance in order to make up for the fact that the wavelet coefficients
are not exactly sparse, as assumed by the BG signal model.
(We note that the same was done in the BG-based work 
\cite{HeCarin:SPL:10, HeCarin:TSP:09}.)
For Turbo-GM, 
the hyperparameters $(a\jL,b\jL)$ for the signal precisions 
$\{\sigma^{-2}\nL\}_{n\in\mc{W}_j}$ were set at the values of $(a_j,b_j)$ 
for the BG case, while the hyperparameters $(a\jS,b\jS)$ for 
$\{\sigma^{-2}\nS\}_{n\in\mc{W}_j}$ were set as $a\jS\!=\!1$ and 
$b\jS=10^{-6}b\jL$.
Meanwhile, the noise variance $\sigma^2$ was assumed to be exactly zero,
because the GM signal prior is capable of modeling non-sparse wavelet
coefficients.

For MCMC \cite{HeCarin:TSP:09}, the hyperparameters were set 
in accordance with the values described in~\cite{HeCarin:TSP:09}; 
the values of $c,d,\{c_j,d_j\}$ are same as the ones used for the proposed
Turbo-BG scheme, while $a=a_j=b=b_j=10^{-6}~\forall j$.
For VB, the same hyperparameters as MCMC were used except for
$a_j=2 \times 10^9$ and $b_j = 10^{10}~\forall j$, which were 
the default values of hyperparameters used in the publicly available 
code.\footnote{\textsl{\url{http://people.ee.duke.edu/~lcarin/BCS.html}}} 
We experimented with the values for both MCMC and VB and found that 
the default values indeed seem to work best. 
For example, if one swaps the $(a_j,b_j)$ hyperparameters between VB 
and MCMC, then the average performance of VB and MCMC 
degrade by $1.69$dB and $1.55$dB, respectively, relative to
the values reported in Table~\ref{tab:summary}.
 
For both the CoSaMP and ModelCS algorithms, the principal tuning parameter is 
the assumed number of non-zero coefficients.
For both ModelCS (which is based on CoSaMP) and CoSaMP itself, we used the
Rice University codes,\footnote{\textsl{\url{http://dsp.rice.edu/software/model-based-compressive-sensing-toolbox}}} which include
a genie-aided mechanism to compute the number of active
coefficients from the original image. 
However, since we observed that the algorithms perform somewhat poorly 
under that tuning mechanism, we instead 
ran (for each image) multiple reconstructions with the number 
of active coefficients varying from $200$ to $2000$ in steps of $100$, and 
reported the result with the best NMSE. 
The number of active coefficients chosen in this manner was usually much 
smaller than that chosen by the genie-aided mechanism.
 
To implement BG-AMP, we used the AMP scheme described in 
Section~\ref{sec:MP:SSR} with the hyperparameter learning scheme described
in Section~\ref{sec:update}; HMT structure was not exploited. 
For this, we assumed that the priors on variance $\sigma_n^2$ and activity 
$\lambda_n$ were identical over the coefficient index $n$, and 
assigned Gamma and Beta hyperpriors of $(a,b)=(10^{-10},10^{-10})$ 
and $(c,d)=(0.1N,0.9N)$, respectively.

For HMT+IRWL1, we ran code provided by the authors with default settings.
For SPGL1,\footnote{\textsl{\url{http://www.cs.ubc.ca/labs/scl/spgl1/index.html}}} 
the residual variance was set to $0$, and all parameters were set at their
defaults.
%

\subsection{Results}

\begin{figure*}[t]
\begin{center}
        \psfrag{original}[B][b][0.8]{\sf Original}
        \psfrag{IRWLS1}[B][b][0.8]{\sf HMT+IRWL1}
        \psfrag{cosamp}[B][b][0.8]{\sf CoSaMP}
        \psfrag{model}[B][b][0.8]{\sf ModelCS}
        \psfrag{bg-amp}[B][b][0.8]{\sf BG-AMP}
        \psfrag{spgl1}[B][b][0.8]{\sf SPGL1}
        \psfrag{vb}[B][b][0.8]{\sf Variational Bayes}
        \psfrag{mcmc}[B][b][0.8]{\sf MCMC}
        \psfrag{turbo}[B][b][0.8]{\sf Turbo-BG}
        \psfrag{turbo-gmm}[B][b][0.8]{\sf Turbo-GM}
\begin{minipage}[b]{0.19\linewidth}
\centering
\includegraphics[width=\columnwidth]{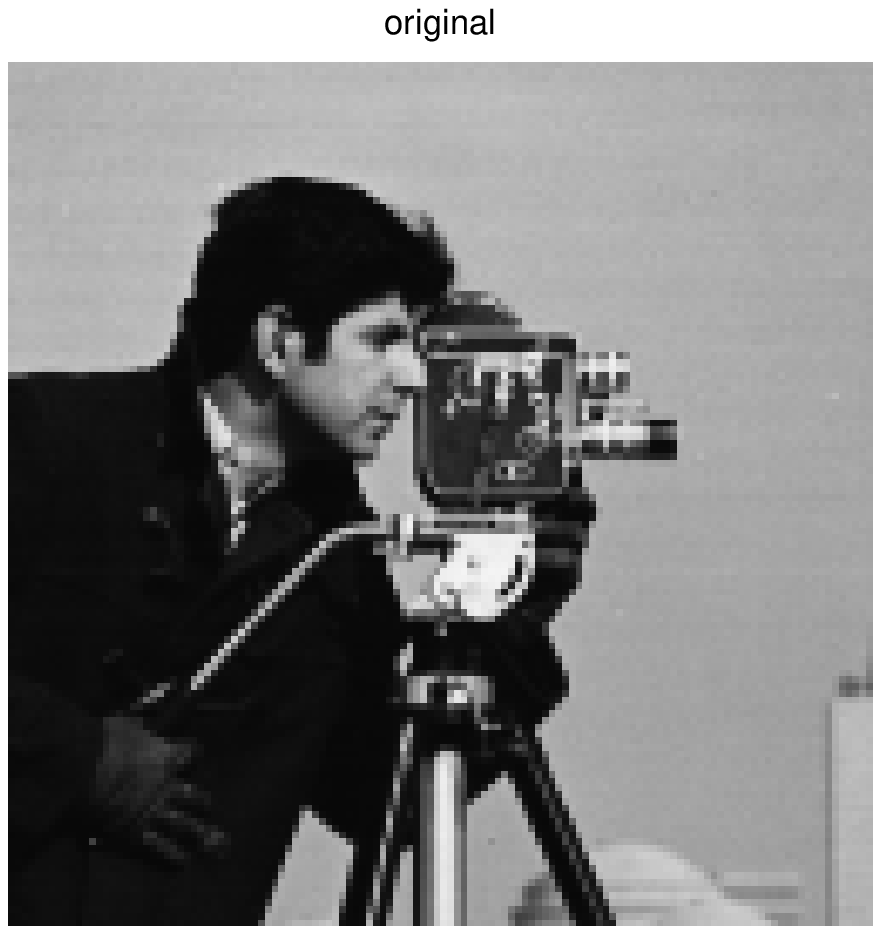}
\end{minipage}
\begin{minipage}[b]{0.19\linewidth}
\centering
\includegraphics[width=\columnwidth]{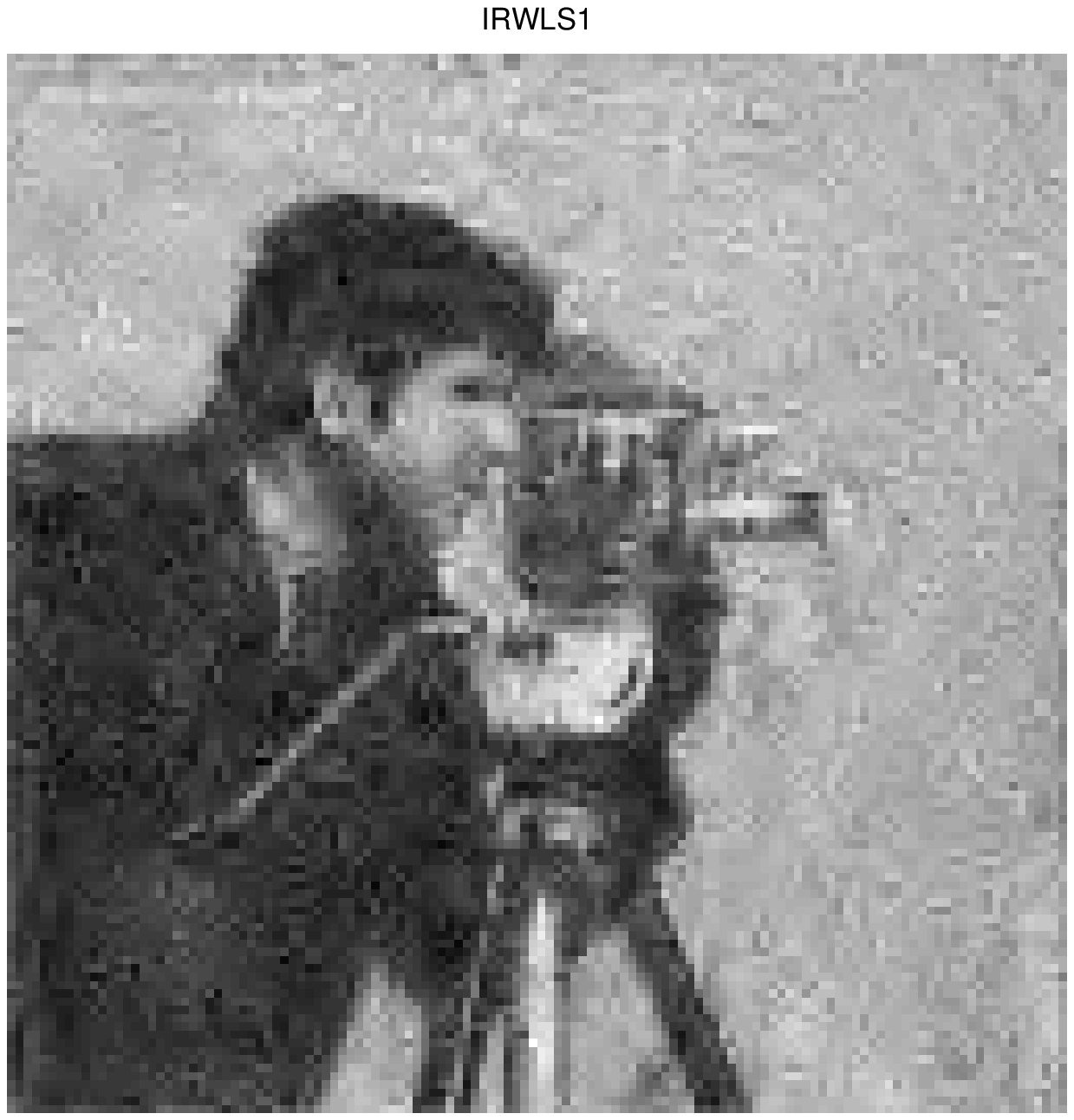}
\end{minipage}
\begin{minipage}[b]{0.19\linewidth}
\centering
\includegraphics[width=\columnwidth]{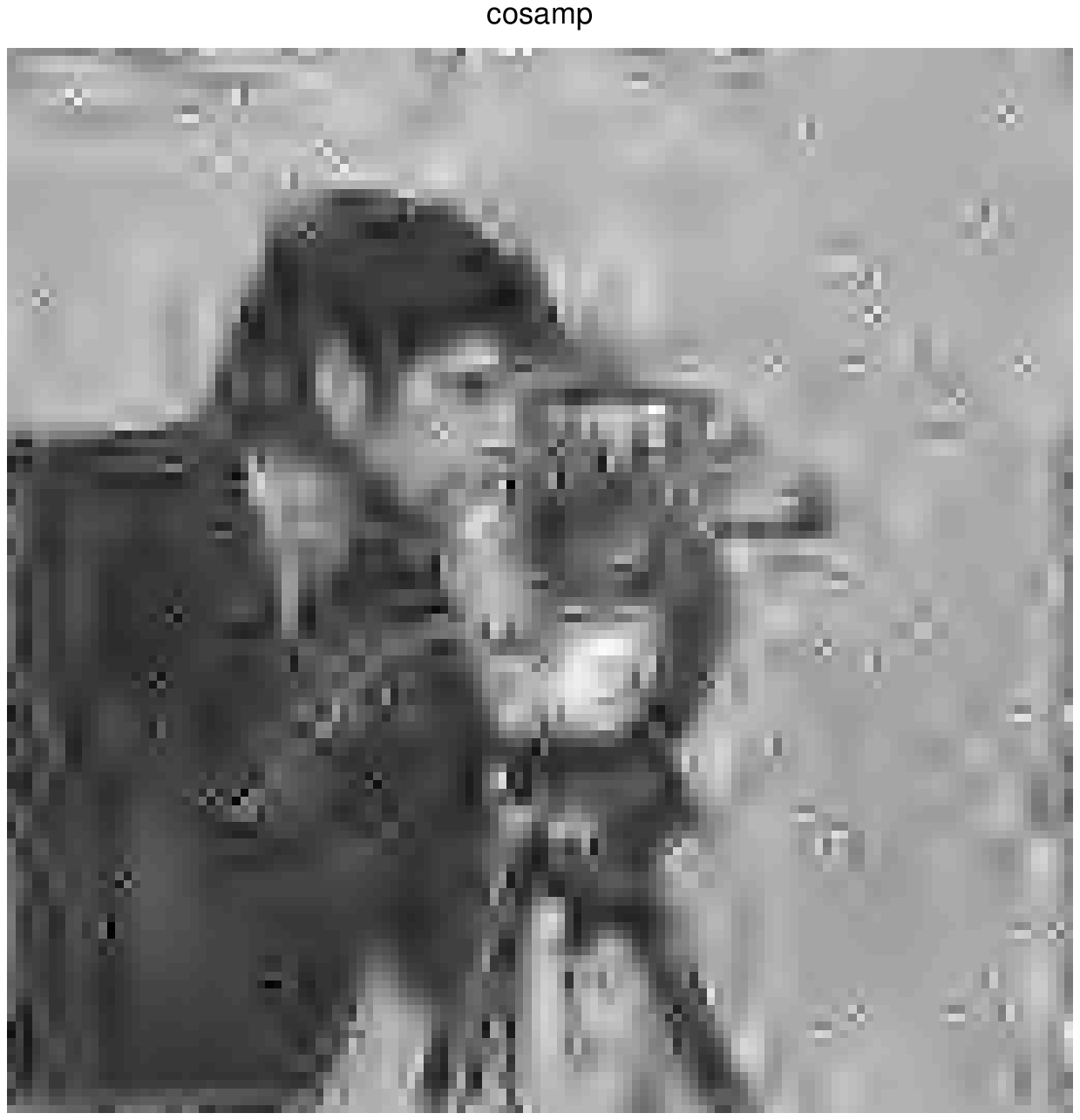}
\end{minipage}
\begin{minipage}[b]{0.19\linewidth}
\centering
\includegraphics[width=\columnwidth]{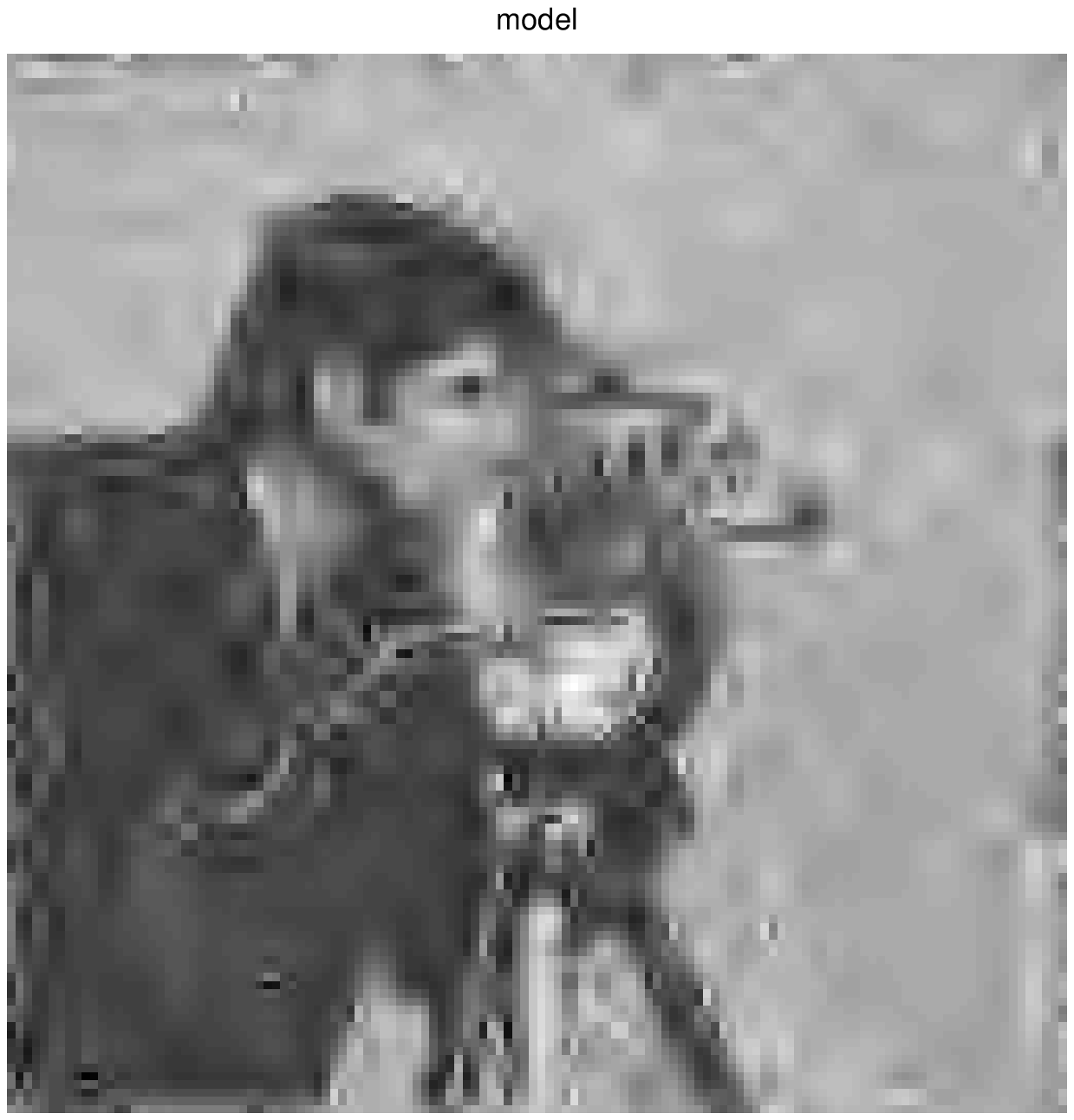}
\end{minipage}
\begin{minipage}[b]{0.19\linewidth}
\centering
\includegraphics[width=\columnwidth]{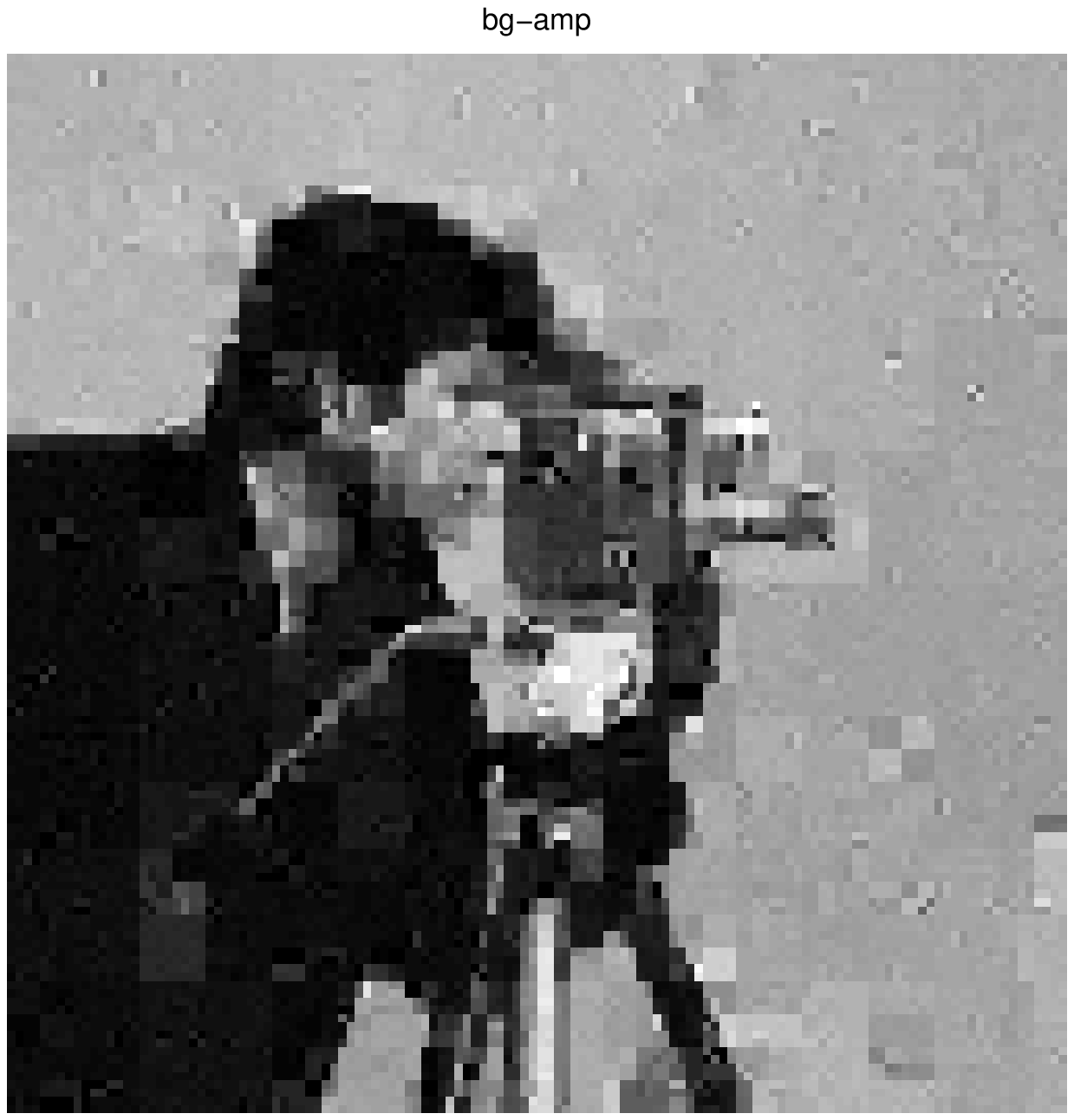}
\end{minipage}
\hspace{0.1mm}\mbox{}\\
\begin{minipage}[b]{0.19\linewidth}
\centering
\includegraphics[width=\columnwidth]{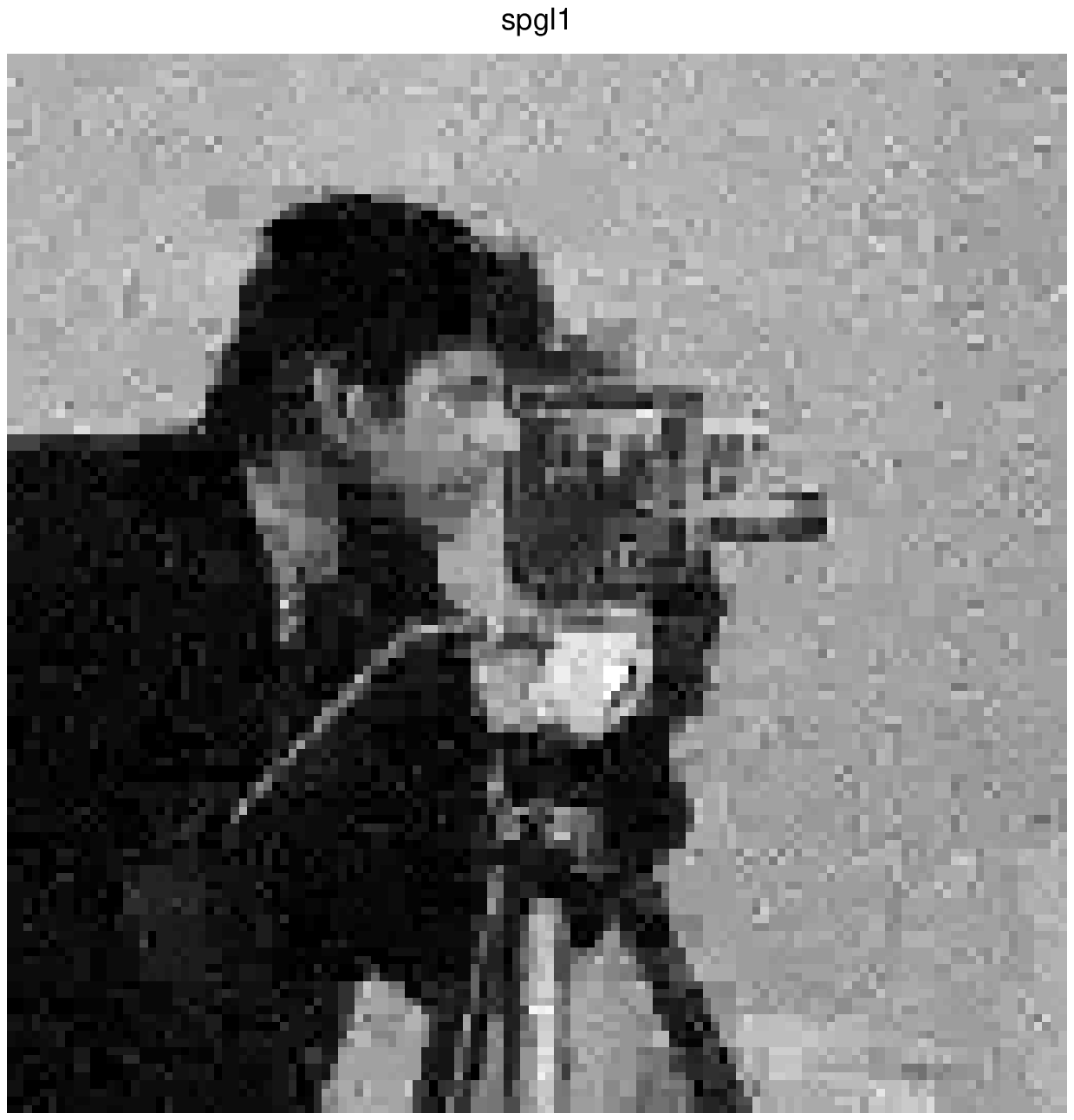}
\end{minipage}
\begin{minipage}[b]{0.19\linewidth}
\centering
\includegraphics[width=\columnwidth]{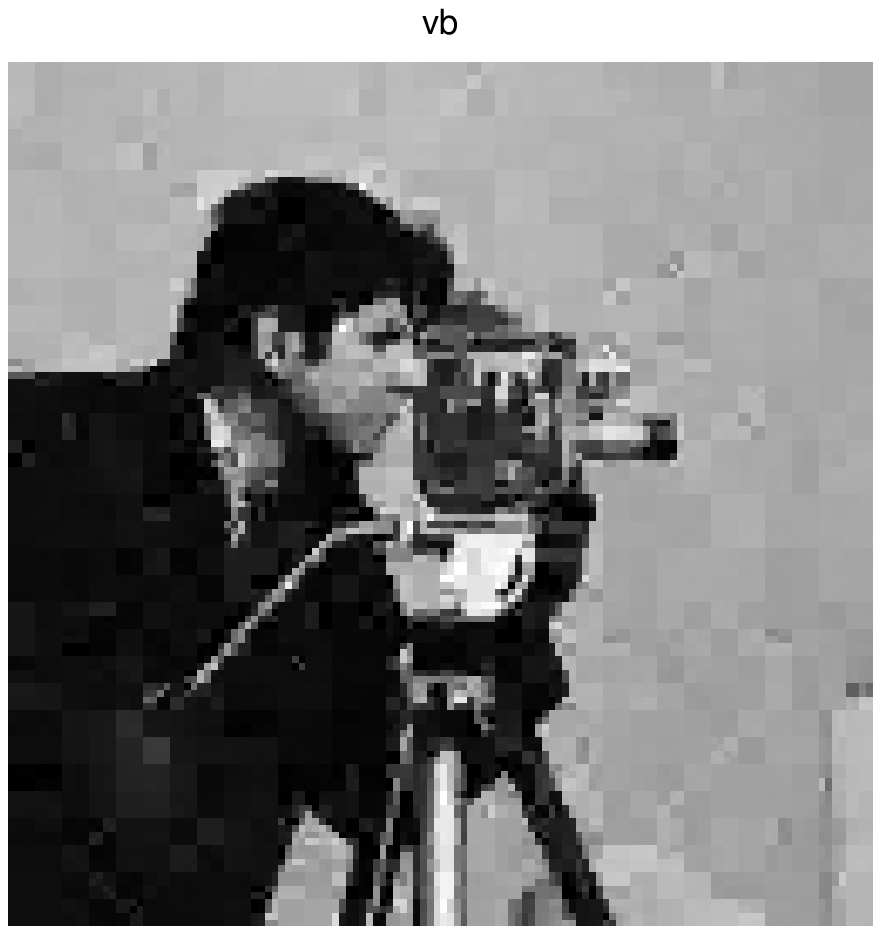}
\end{minipage}
\begin{minipage}[b]{0.19\linewidth}
\centering
\includegraphics[width=\columnwidth]{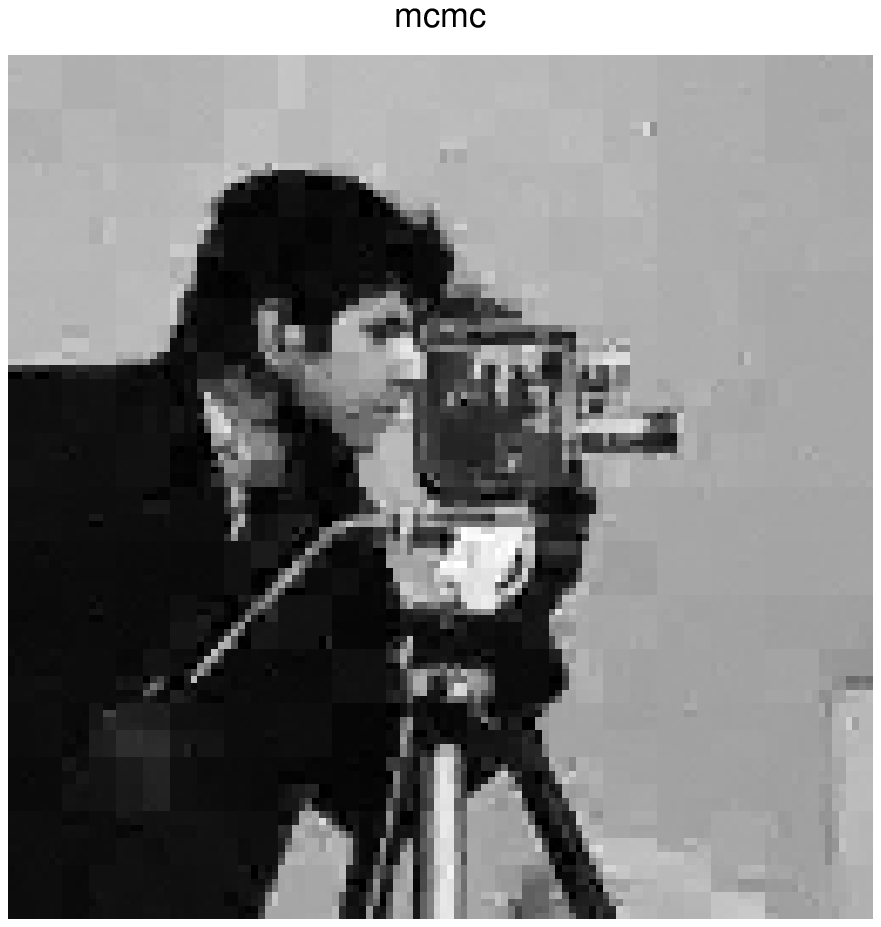}
\end{minipage}
\begin{minipage}[b]{0.19\linewidth}
\centering
\includegraphics[width=\columnwidth]{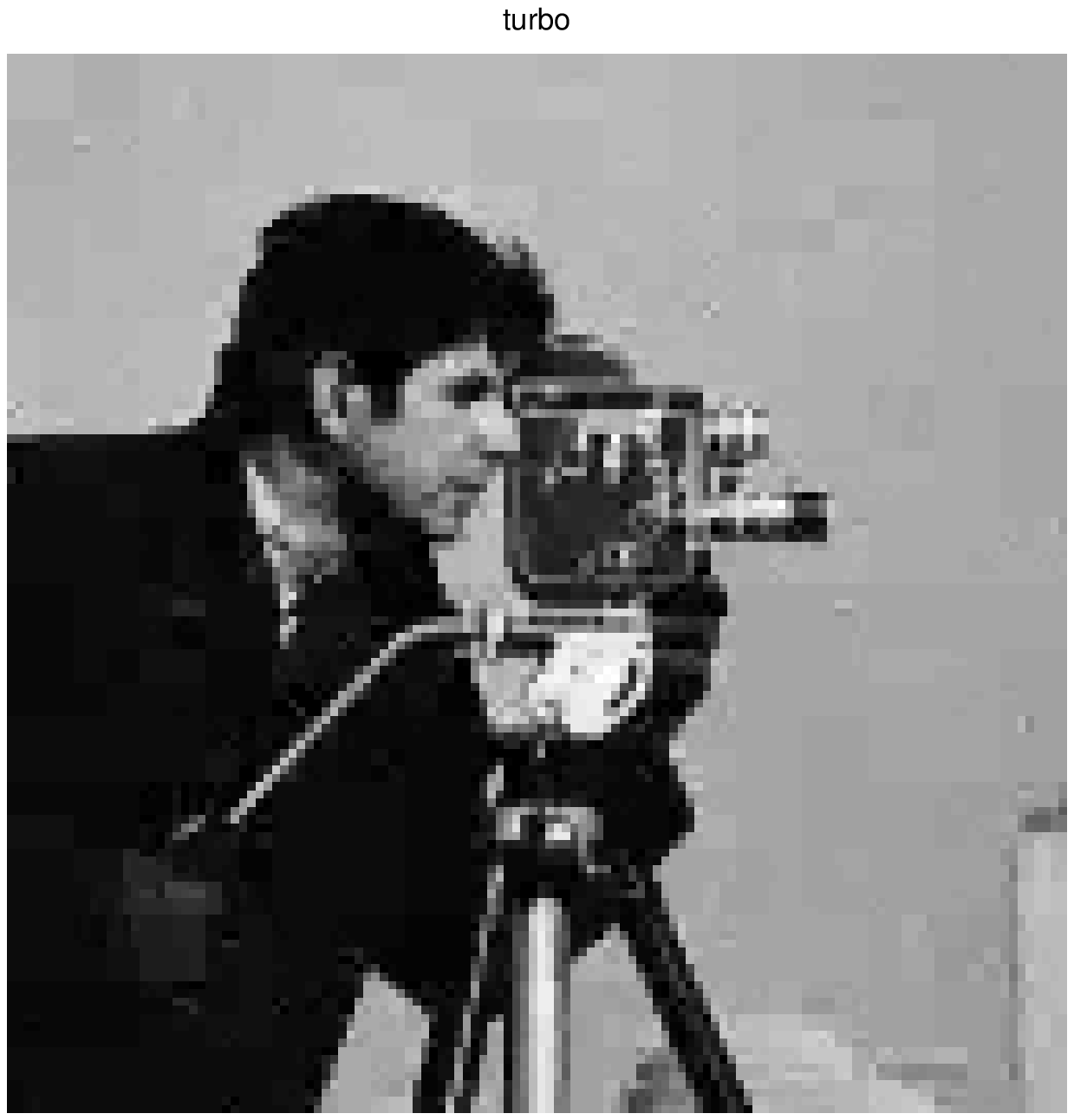}
\end{minipage}
\begin{minipage}[b]{0.19\linewidth}
\centering
\includegraphics[width=\columnwidth]{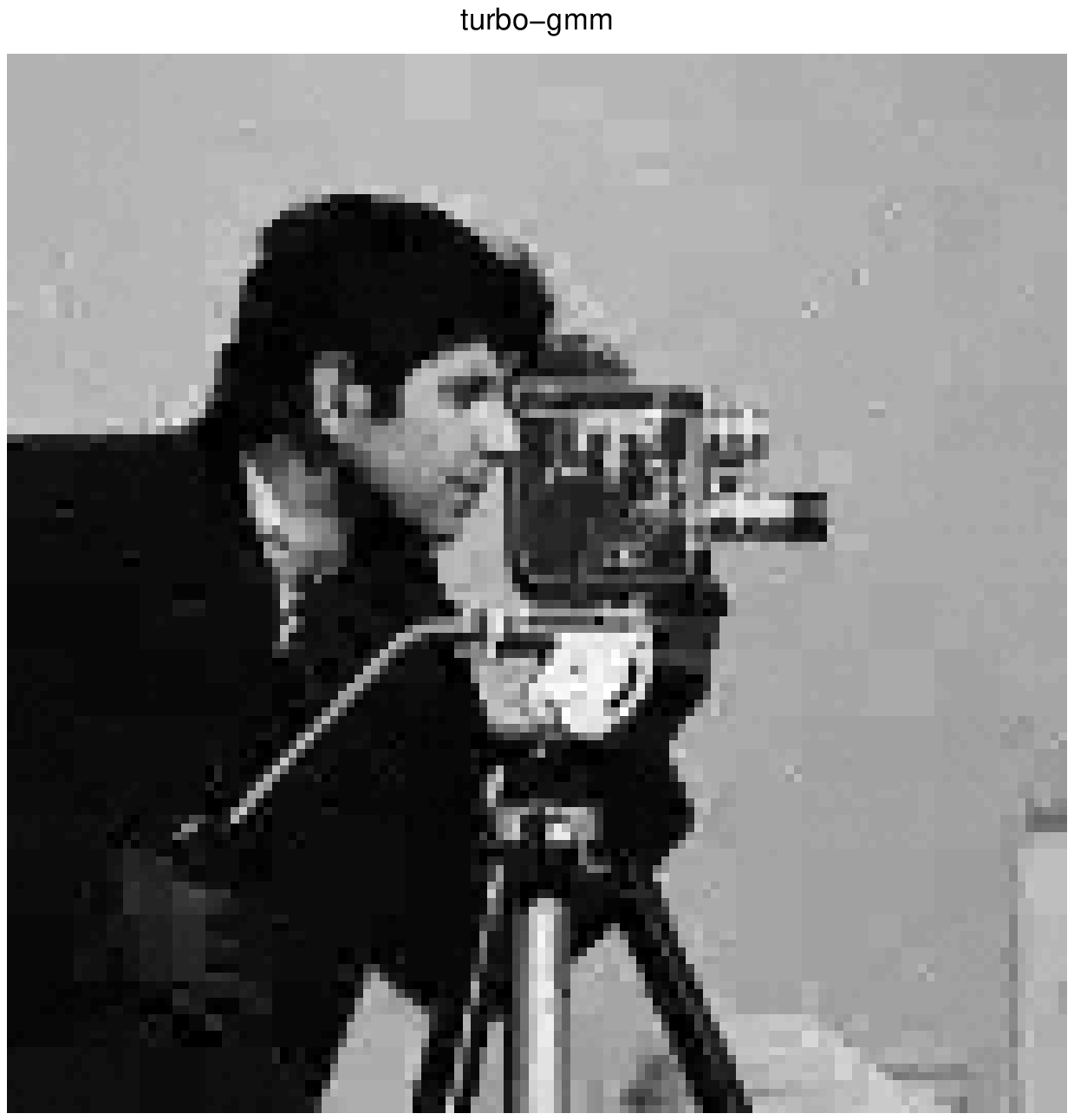}
\end{minipage}
\caption{Reconstruction from $M=5000$ observations of a $128 \times 128$ (i.e., $N=16384$) section
of the cameraman image using i.i.d Gaussian $\vec{\Phi}$.}
\label{fig:recComp}
\end{center}
\end{figure*}

Fig.~\ref{fig:recComp} shows a $128\! \times\! 128$ section of the 
``cameraman'' image along with the images recovered by the various algorithms.
Qualitatively, we see that CoSaMP, which leverages only simple sparsity, 
and ModelCS, which models persistence-across-scales (PAS) through a
deterministic tree structure, both perform relatively poorly. 
HMT+IRWL1 also performs relatively poorly, due to (we believe) the
ad-hoc manner in which the HMT structure was exploited via
iteratively re-weighted $\ell_1$.
The BG-AMP and SPGL1 algorithms, neither of which attempt to exploit 
PAS, perform better. 
The HMT-based schemes (VB, MCMC, Turbo-GM, and Turbo-GM) all perform 
significantly better, with the Turbo schemes performing best. 

\begin{figure}[t]
\begin{center}
        \epsfig{file=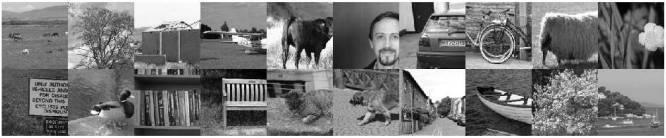,width=\linewidth,clip=} 
\end{center}
\vspace{-2mm}
\caption{A sample image from each of the 20 types in the Microsoft database. 
Image statistics were found to vary significantly from one type to another.}
\label{fig:categories}
\end{figure}

    \begin{figure}[t]
    \begin{center}
        \psfrag{x}[c][Bc][0.8]{\sf Image type}
        \psfrag{y}[c][c][0.8]{\sf Average NMSE (dB)}
        \psfrag{RiceCoSaMP}[l][l][0.45]{\sf \hspace{-1.3mm} CoSaMP}
        \psfrag{IRWL1}[l][l][0.45]{\sf \hspace{-1.3mm} HMT-IRWL1}
        \psfrag{Turbo}[l][l][0.45]{\sf Turbo--BG}
        \psfrag{Turbo-GMM}[l][l][0.45]{\sf Turbo--GM}
        \epsfig{file=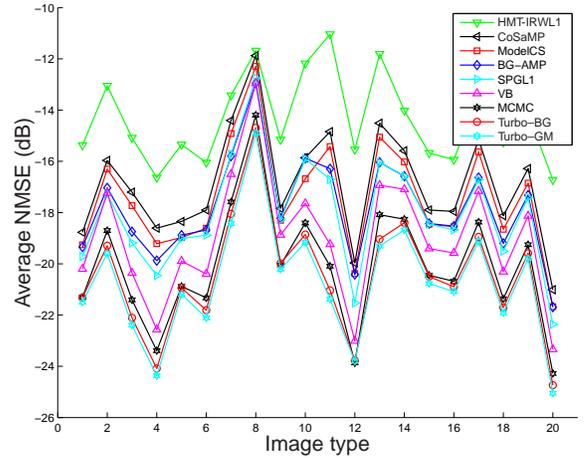,width=3.0in,clip=} 
    \end{center}
    \caption{Average NMSE for each image type.}
    \label{fig:NMSE}
    \end{figure}
    
    \begin{figure}[t]
    \begin{center}
        \psfrag{x}[c][Bc][0.8]{\sf Image type}
        \psfrag{y}[c][c][0.8]{\sf Average runtime (sec)}
        \psfrag{RiceCoSaMP}[l][l][0.45]{\sf \hspace{-1.3mm} CoSaMP}
        \psfrag{IRWL1}[l][l][0.45]{\sf \hspace{-1.3mm} HMT-IRWL1}
        \psfrag{Turbo}[l][l][0.45]{\sf Turbo--BG}
        \psfrag{Turbo-GMM}[l][l][0.45]{\sf Turbo--GM}
        \epsfig{file=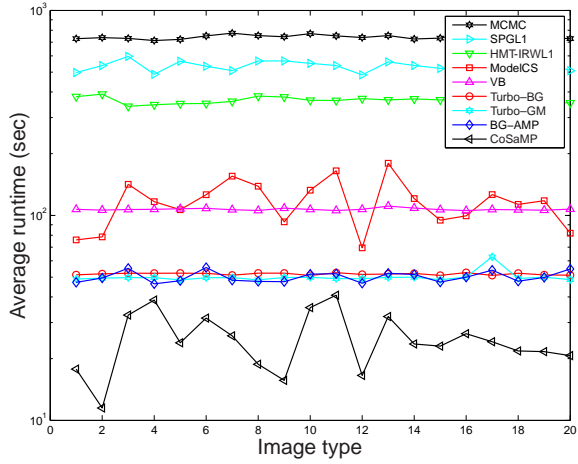,width=3.0in,clip=} 
    \end{center}
    \caption{Average runtime for each image type.} 
    \label{fig:compTime}
    \end{figure}

\begin{table}[t]
\centering
\begin{tabular}{|c||c|c|}\hline
Algorithm & NMSE (dB)  & Computation Time (sec)\\
\hline \hline
HMT+IRWL1   &  -14.37  &     363 \\
\hline 
CoSaMP          &     -16.90  &                 25\\
\hline 
ModelCS             &        -17.45    &               117\\
\hline 
BG-AMP              &       -17.84     &               68\\
\hline 
SPGL1               &         -18.06   &                536\\
\hline 
VB                  &            -19.04 &                  107\\
\hline 
MCMC                &         -20.10    &              742\\
\hline 
Turbo-BG           &            -20.49 &                 51\\
\hline 
Turbo-GM           &       -20.74      &            51  \\
\hline 
\end{tabular}
\caption{NMSE and runtime averaged over $591$ images.}
\label{tab:summary}
\end{table}

    \begin{figure}[t]
    \begin{center}
        \psfrag{x}[c][Bc][0.8]{\sf Number of measurements $(M)$}
        \psfrag{y}[c][c][0.8]{\sf Average NMSE (dB)}
        \psfrag{RiceCoSaMP}[l][l][0.45]{\sf \hspace{-1.3mm} CoSaMP}
        \psfrag{IRWL1}[l][l][0.45]{\sf \hspace{-1.3mm} HMT-IRWL1}
        \psfrag{Turbo}[l][l][0.45]{\sf Turbo--BG}
        \psfrag{Turbo-GMM}[l][l][0.45]{\sf Turbo--GM}
        \epsfig{file=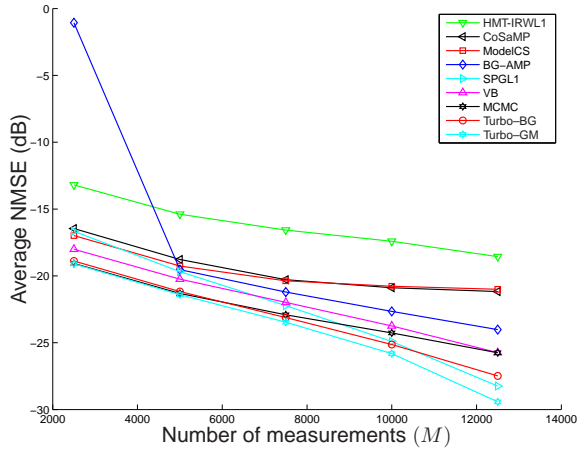,width=3.0in,clip=} 
    \end{center}
    \caption{Average NMSE for images of type 1.}
    \label{fig:NMSE_varyM}
    \end{figure}
    
    \begin{figure}[t]
    \begin{center}
        \psfrag{x}[c][Bc][0.8]{\sf Number of measurements $(M)$}
        \psfrag{y}[c][c][0.8]{\sf Average runtime (sec)}
        \psfrag{RiceCoSaMP}[l][l][0.45]{\sf \hspace{-1.3mm} CoSaMP}
        \psfrag{IRWL1}[l][l][0.45]{\sf \hspace{-1.3mm} HMT-IRWL1}
        \psfrag{Turbo}[l][l][0.45]{\sf Turbo--BG}
        \psfrag{Turbo-GMM}[l][l][0.45]{\sf Turbo--GM}
        \epsfig{file=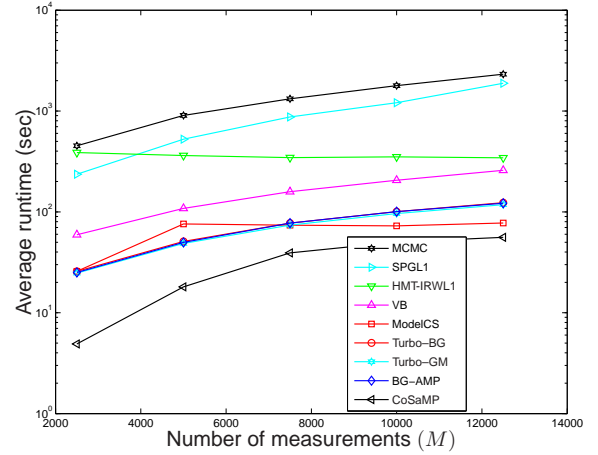,width=3.0in,clip=} 
    \end{center}
    \caption{Average runtime for images of type 1.} 
    \label{fig:compTime_varyM}
    \end{figure}


For a quantitative comparison, we measured average performance over a suite 
of images in a \emph{Microsoft Research Object Class Recognition} 
database\footnote{
 We used $128\!\times\!128$ images extracted from the 
 ``Pixel-wise labelled image database v2'' at
 \textsl{\url{http://research.microsoft.com/en-us/projects/objectclassrecognition}}. 
 What we refer to as an ``image type'' is a ``row'' in this database.
}
that contains $20$ types of images (see Fig.~\ref{fig:categories}) with roughly
$30$ images of each type. 
In particular, we computed the average NMSE and
average runtime on a 2.5~GHz PC, for each image type. 
These results are reported in Figures~\ref{fig:NMSE} and~\ref{fig:compTime},
and the global averages (over all $591$ images) are reported in 
Table~\ref{tab:summary}. 
From the table, we observe that the proposed Turbo algorithms outperform 
all the other tested algorithms in terms of reconstruction NMSE, but are
beaten only by CoSaMP in speed.\footnote{
  The CoSaMP runtimes must be interpreted with caution, because the 
  reported runtimes correspond to a single reconstruction, whereas 
  in practice multiple reconstructions may be needed to determined the
  best value of the tuning parameter.}
Between the two Turbo algorithms, we observe that Turbo-GM slightly 
outperforms Turbo-BG in terms of reconstruction NMSE, while taking the 
same runtime.
In terms of NMSE performance, the closest competitor to the Turbo schemes 
is MCMC,\footnote{
  The MCMC results reported here are for the defaults settings: 
  100 MCMC iterations and 200 burn-in iterations.
  Using 500 MCMC iterations and 200 burn-in iterations, we obtained
  an average NMSE of $-20.22$dB (i.e., $0.12$dB better) at
  an average runtime of $1958$ sec (i.e., $2.6\times$ slower).}
whose NMSE is $0.39$dB worse than Turbo-BG and $0.65$dB worse 
than Turbo-GM. 
The good NMSE performance of MCMC comes at the cost of complexity, though: 
MCMC is $15$ times slower than the Turbo schemes.
The second closest NMSE-competitor is VB, showing performance $1.5$~dB 
worse than Turbo-BG and $1.7$dB worse than Turbo-GM.
Even with this sacrifice in performance, VB is still twice as slow as 
the Turbo schemes.
Among the algorithms that do not exploit PAS, we see that 
SPGL1 offers the best NMSE performance, but is by far the slowest
(e.g., $20$ times slower than CoSaMP).
Meanwhile, CoSaMP is the fastest, but shows the worst NMSE performance
(e.g., $1.16$dB worse than SPGL1).
BG-AMP strikes an excellent balance between the two: its NMSE 
is only $0.22$dB away from SPGL1, whereas it takes only $2.7$ times as long
as CoSaMP.
%
However, by combining the AMP algorithm with HMT structure via the
turbo approach, it is possible to significantly improve NMSE while 
simultaneously decreasing the runtime. 
The reason for the complexity decrease is twofold.
First, the HMT structure helps the AMP and parameter-learning iterations 
to converge faster.
Second, the HMT steps are computationally negligible relative to the AMP
steps: when, e.g., $M=5000$, the AMP portion of the turbo iteration takes 
approximately $6$ sec while the HMT portion takes $0.02$ sec.

We also studied NMSE and compute time as a function of the number of 
measurements, $M$. 
For this study, we examined images of Type~1 at 
$M=2500, 5000, 7500, 10000, 12500$.
In Figure~\ref{fig:NMSE_varyM}, we see that Turbo-GM offers the uniformly
best NMSE performance across $M$.
However, as $M$ decreases, there is little difference between the NMSEs 
of Turbo-GM, Turbo-CS, and MCMC.
As $M$ increases, though, we see that the NMSEs of MCMC and VB converge,
but that they are significantly outperformed by Turbo-GM, Turbo-CS, 
and---somewhat surprisingly---SPGL1.
In fact, at $M=12500$, SPGL1 outperforms Turbo-BG, but not Turbo-GM.
However, the excellent performance of SPGL1 at these $M$ comes at the 
cost of very high complexity, as evident in Figure~\ref{fig:compTime_varyM}.
\section{Conclusion}				\label{sec:conc} 
We proposed a new approach to HMT-based compressive imaging based on loopy
belief propagation, leveraging a turbo message passing schedule and 
the AMP algorithm of Donoho, Maleki, and Montanari.
We then tested our algorithm on a suite of $591$ natural images and found that
it outperformed the state-of-the-art approach (i.e., variational Bayes) 
while halving its runtime.
\bibliographystyle{IEEEtran}

\end{document}